\documentclass{article}

\usepackage[preprint]{neurips_2026}
\usepackage{amsmath, amssymb}
\usepackage{booktabs}
\usepackage{siunitx} 
\usepackage{xcolor}
\usepackage{graphicx}
\usepackage{subcaption}
\usepackage{caption}
\usepackage{multirow}
\usepackage{float}
\usepackage{enumitem}

\usepackage[utf8]{inputenc} 
\usepackage[T1]{fontenc}    
\usepackage{hyperref}       
\usepackage{url}            
\usepackage{booktabs}       
\usepackage{amsfonts}       
\usepackage{nicefrac}       
\usepackage{microtype}      
\usepackage{xcolor}         

\title{NeuronEye: Query-Guided Visual Concept Activation for Vision-Language Reasoning}

\author{
Ruiyu Yan$^{1}$ \quad Bowen Chen$^{2}$ \quad Shaowen Wan$^{2}$ \quad Lin Zhao$^{2}$\\
$^{1}$New York University \quad
$^{2}$New Jersey Institute of Technology\\
\texttt{lin.zhao.1@njit.edu}
}

\begin{document}

\maketitle

\begin{abstract}
Current vision-language models (VLMs) encode visual information in dense hidden states where object identity, spatial layout, and local attributes are implicitly entangled rather than explicitly disentangled, limiting their ability to isolate and modulate the specific visual evidence required by a given language query. Inspired by sparse population coding and top-down modulation in biological vision, we introduce \emph{NeuronEye}, a plug-in framework that constructs a sparse, concept-level neuron vocabulary from intermediate VLM representations and selectively activates query-relevant visual concepts during inference. NeuronEye decomposes vision-token states into an overcomplete sparse basis organized by concept-level clusters, uses the language query to activate relevant clusters and localize the patches where selected concepts are expressed, and injects the focused evidence back into vision tokens. A complementary suppression mechanism attenuates dominant perceptual directions to preserve weaker but relevant cues. All operations run in a single forward pass over a frozen VLM backbone. On Qwen2.5-VL-7B, NeuronEye raises CV-Bench overall accuracy by +3.1 with gains of +9.5 on Distance, and improves BLINK Multi-view by +8.3, with similar trends on LLaVA-1.6-7B. These results suggest that sparse neuron vocabularies can serve not only as post-hoc interpretability tools but also as active interfaces for concept-level visual reasoning.

\vspace{-1em}
\begin{figure}[htbp]
    \centering
    \includegraphics[width=0.88\linewidth]{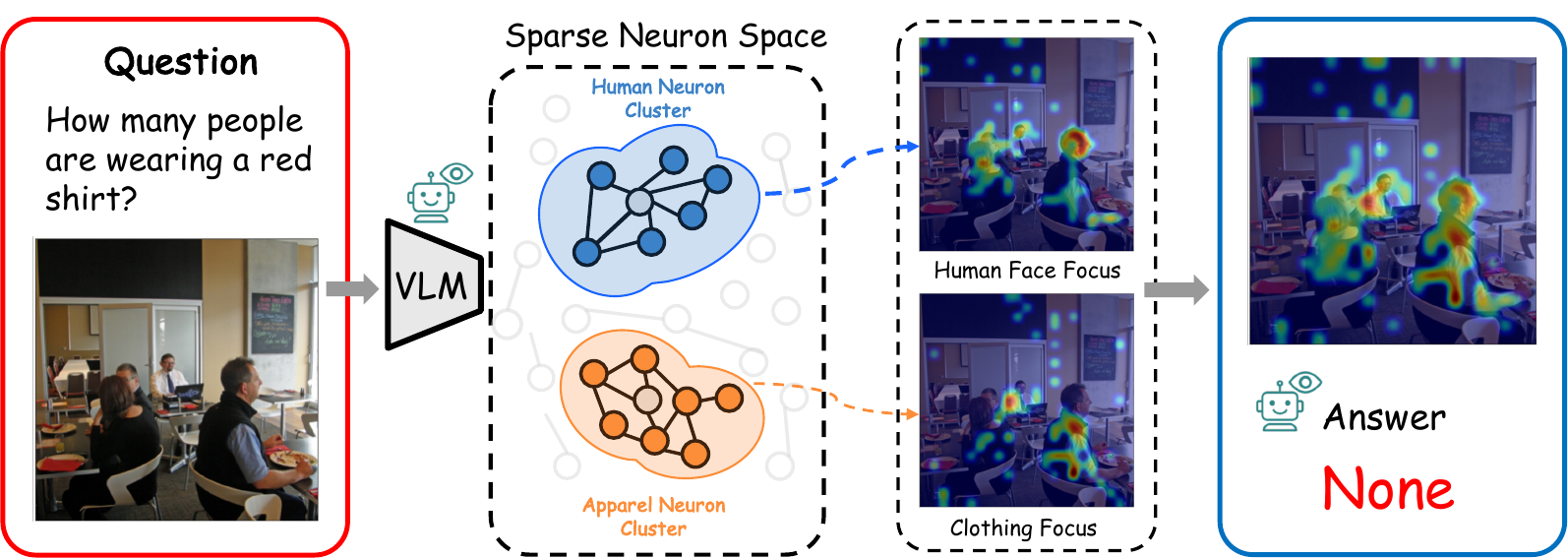}
\caption{
\textbf{NeuronEye overview.}
  Given an image and a language query, NeuronEye decomposes
  intermediate VLM representations into a sparse neuron vocabulary
  and identifies query-relevant concept clusters (e.g., \emph{Human}
  and \emph{Apparel}). Each cluster localizes its corresponding
  visual evidence in the image, and the focused activations are
  injected back into the VLM to guide reasoning.
}
    \label{fig:abstract}
\end{figure}
\end{abstract}

\section{Introduction}
Modern vision-language models (VLMs) align image and text by jointly processing visual and language tokens through learned attention-based multimodal fusion~\cite{liu2023visual,bai2025qwen25vl,li2024llavaonevision}. Although effective on a wide range of tasks, this paradigm encodes visual information in dense hidden states where object identity, spatial layout, local attributes, and background context are not explicitly disentangled. Recent work has shown that VLMs exhibit multi-object reasoning failures remarkably similar to those caused by representational interference in human rapid feedforward vision~\cite{campbell2024understanding}, and that the bottleneck in spatial reasoning often lies in integrating visual information rather than in perceiving it~\cite{tong2024cambrian1,fu2024blink}. These findings suggest that dense visual representations may be insufficient for isolating and selectively modulating individual visual concepts according to the question at hand.
 
The biological visual system addresses this difficulty through two complementary mechanisms. Neurons in primary visual cortex represent natural scenes via sparse population codes, where any stimulus activates only a small subset of neurons while the majority remain silent~\cite{olshausen1996emergence,vinje2000sparse}. At higher cortical levels, this sparsity becomes even more selective: concept cells in the human medial temporal lobe respond to specific persons or landmarks regardless of viewpoint or input modality~\cite{quiroga2005invariant,quiroga2012concept}, suggesting that the brain organizes visual information into a sparse, concept-level neural vocabulary whose entries can be independently addressed. Then, top-down attention selectively modulates this vocabulary according to task demands. Treisman's feature integration theory established that separately processed visual features require focused attention to be correctly bound into object percepts~\cite{treisman1980feature,treisman1998feature}, and neurophysiological studies have confirmed that prefrontal top-down signals enhance task-relevant neural responses while suppressing competing representations~\cite{desimone1995neural,noudoost2010top}. Together, these mechanisms implement a two-stage strategy: decompose visual input into a sparse concept-level vocabulary, then selectively modulate task-relevant entries to guide downstream perception and reasoning.
 
Inspired by this perspective, we introduce \emph{NeuronEye}, a framework that constructs a sparse, concept-level neuron vocabulary from intermediate VLM representations and uses language queries to selectively activate relevant visual concepts during inference (Fig.~\ref{fig:abstract}). NeuronEye has three core components. \emph{Sparse Neuron Space} (SNS) projects vision-token states into an overcomplete sparse basis via a sparse autoencoder (SAE) and groups the resulting neurons into concept-level clusters, forming an addressable visual concept vocabulary. \emph{Neuron-guided Visual Focus} (NVF) uses the language query, together with image-side evidence, to activate relevant neuron clusters, locate the patches where the selected concepts are expressed, and inject the activated evidence back into the corresponding vision tokens. \emph{Perceptual Concept Suppression} (PCS) complements NVF by attenuating dominant perceptual directions at a later layer, preventing focused activation from suppressing weaker but relevant cues. All operations run in a single forward pass with the VLM backbone and SAE frozen.

We evaluate NeuronEye on four vision-centric benchmarks, demonstrating that it consistently improves structured visual reasoning. Applied to Qwen2.5-VL-7B, NeuronEye raises CV-Bench overall accuracy from 78.5 to 81.6 (+3.1), with gains of +9.5 on Distance and +2.3 on Relation, and improves BLINK Multi-view by +8.3. The method also transfers to LLaVA-1.6-7B with similar trends on spatial and relational sub-tasks. Compared with vision-token reduction and representation-level intervention baselines under the same controlled evaluation protocol, NeuronEye achieves the best overall performance on both CV-Bench and BLINK without additional finetuning.

Our contributions are summarized as follows:

\begin{itemize}
    \item We propose NeuronEye, a concept-level selective modulation framework that decomposes dense visual representations into a sparse neuron vocabulary organized by visual concepts, and activates query-relevant neuron clusters to modulate visual evidence for VLM reasoning. 
    \item NeuronEye operates as a plug-in module over a frozen VLM backbone. It requires only a one-time sparse autoencoder training on intermediate vision-token activations, and all inference is completed in a single forward pass.
    \item We show that sparse autoencoder features, which have been used exclusively as a post-hoc interpretability tool, can serve as a structured vocabulary for concept-level visual reasoning, extending their role from passive interpretation to actively improving model performance.
    \item We evaluate NeuronEye on four vision-centric benchmarks across two VLM backbones, demonstrating consistent improvements on spatial and localization-sensitive tasks.
    
\end{itemize}

\section{Related Works}

\subsection{Visual Representation Modulation}
Prior work has explored visual representation modulation to focus VLM reasoning on relevant visual evidence. Methods such as FastV~\cite{chen2024fastv}, SparseVLM~\cite{zhang2025sparsevlm}, PruMerge~\cite{shang2025prumerge}, and PyramidDrop~\cite{xing2025conical} remove, merge, or retain patch tokens based on attention or importance scores, determining \emph{where} the model attends. However, each retained token remains a dense unit carrying all visual concepts together; selecting a token retains all its encoded information, including information irrelevant to the query. Representation-level methods such as VISTA~\cite{li2025vista} adjust \emph{what} is represented by steering dense hidden states, but these modifications are applied uniformly and do not select which visual concept to enhance for a given query. NeuronEye operates at a finer granularity. Rather than selecting which patch tokens to keep or uniformly steering dense hidden states, it decomposes vision-token representations into a sparse concept-level neuron vocabulary and uses the language query to activate relevant concept subsets within localized patches for targeted modulation.

\subsection{Sparse Autoencoders in Multimodal Models}
Sparse autoencoders (SAEs) have been widely adopted to decompose superposed activations into sparse, interpretable latent features~\cite{cunningham2024sparse, pach2025sparse}. In this post-hoc setting, SAE features serve as a lens for understanding model internals, revealing human-interpretable directions that correspond to recognizable concepts in the representation space~\cite{bricken2023monosemanticity,zhang2025largemultimodalmodelsinterpret}. More recently, several works have moved beyond interpretation toward active intervention, using SAE features to steer model behavior. By amplifying or suppressing selected latent directions, these methods bias model outputs toward desired content or away from undesired behavior~\cite{zou2024representation, rimsky2024steering}. In the multimodal setting, SAVE~\cite{park2026save} and SSL~\cite{hua2025ssl} apply this steering paradigm to VLMs, using SAE features to identify hallucination-related directions and globally suppress them during generation. While effective for hallucination mitigation, these methods identify a fixed set of SAE features offline and apply them without conditioning on the specific query, and do not organize features into structured groups for selective routing. They therefore do not target query-specific visual reasoning or fine-grained evidence selection. NeuronEye goes beyond steering by organizing SAE-derived visual features into concept-level neuron clusters, selecting relevant clusters conditioned on the language query, and activating them within localized image patches, turning the sparse feature basis into a structured vocabulary for visual reasoning.

\section{Method}

NeuronEye operates through three stages. Given an image--question pair, \emph{Sparse Neuron Space} (SNS, Section~\ref{sec:sparse_representation}) first decomposes intermediate visual representations into an overcomplete sparse basis and organizes the resulting neurons into concept-level clusters, forming an addressable visual concept vocabulary. \emph{Neuron-guided Visual Focus} (NVF, Section~\ref{sec:visual_focus}) then uses the language query and image-side evidence to activate relevant neuron clusters, localize the patches where selected concepts are expressed, and inject the activated evidence back into the corresponding vision tokens at layer $\ell$. Finally, \emph{Perceptual Concept Suppression} (PCS, Section~\ref{sec:pcs}) attenuates dominant perceptual directions at a later layer $\ell'$ to preserve representational diversity after focused activation.

\subsection{Sparse Neuron Space (SNS)}
\label{sec:sparse_representation}
\begin{figure}[htbp]
    \centering
    \includegraphics[width=1\linewidth]{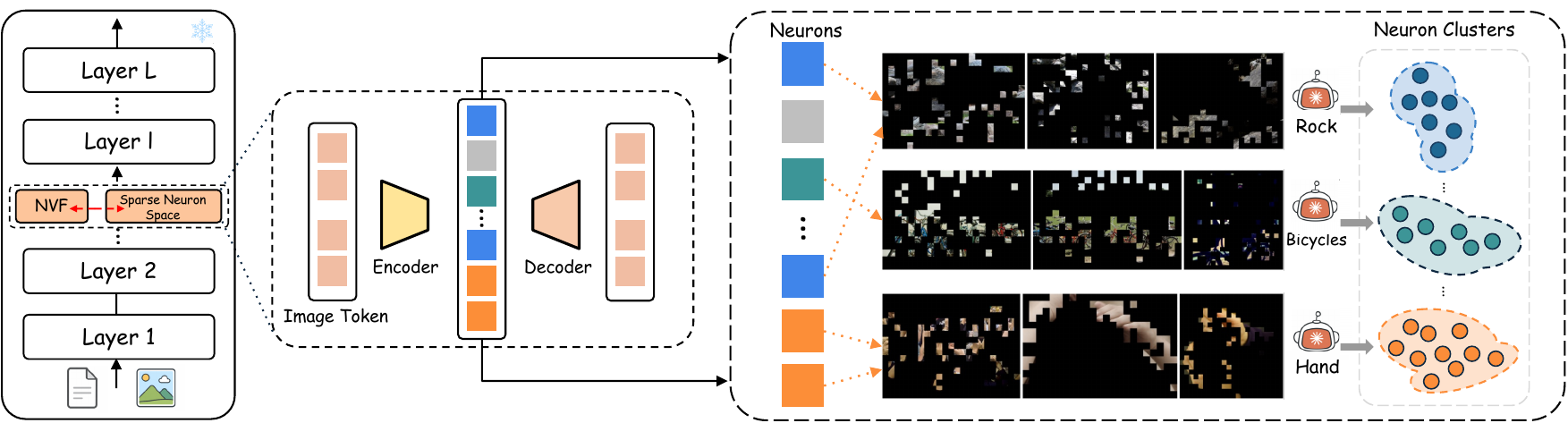}
    \caption{Construction of Sparse Neuron Space. Intermediate vision-token states are encoded by a sparse autoencoder, whose activated neurons are visualized through top-activating patches and grouped into concept-level neuron clusters.
    }
    \label{fig:sparse_neuron_space}
\end{figure}

\paragraph{Sparse Neuron Extraction.}
Given an image--question pair, we pass the visual input and textual query through the frozen VLM and extract the hidden representation of each image patch token at a designated intermediate layer $\ell$, denoted as $\mathbf{x}_{i,p} \in \mathbb{R}^d$, where $i$ indexes the image and $p$ indexes the patch token. We use a SAE to decompose each patch representation into a set of neuron activations (Fig.~\ref{fig:sparse_neuron_space}): the SAE projects $\mathbf{x}_{i,p}$ into an overcomplete latent space $\mathbb{R}^D$ with $D = \alpha d$ ($\alpha \gg 1$) using a linear encoder followed by ReLU activation, where each latent dimension corresponds to a \emph{neuron}. A deterministic Top-$k$ operator then retains only the $k$ largest activations and zeros out the rest, producing a sparse code $\mathbf{z}_{i,p}$ with fixed $\ell_0 = k$ sparsity:
\begin{equation}
\mathbf{z}_{i,p} =
\mathrm{Top}_k\!\bigl(\mathrm{ReLU}(\mathbf{W}_{\mathrm{enc}}\,\mathbf{x}_{i,p})\bigr).
\end{equation}
The representation is thus sparse because only $k$ out of $D$ neurons are active for any given patch, and each active neuron carries a scalar activation indicating how strongly that visual concept is expressed at that patch location. The patch representation is reconstructed as $\hat{\mathbf{x}}_{i,p} = \mathbf{W}_{\mathrm{dec}}\,\mathbf{z}_{i,p}$, where decoder columns are $\ell_2$-normalized to prevent scale degeneracy. Training minimizes a patch-level reconstruction objective augmented by an $\ell_1$ activation penalty:
\begin{equation}
\mathcal{L}_{\mathrm{SAE}} =
\frac{1}{|\mathcal{B}|}
\sum_{(i,p)\in \mathcal{B}}
\left\|
\mathbf{x}_{i,p} - \hat{\mathbf{x}}_{i,p}
\right\|_2^2
+
\lambda_s
\left\|
\mathbf{z}_{i,p}
\right\|_1 ,
\end{equation}
where $\mathcal{B}$ denotes a minibatch of image patch tokens. Since the Top-$k$ operator fixes the number of active coordinates, the $\ell_1$ term mainly regularizes the magnitude of selected activations.

\paragraph{Neuron Filtering and Neuron Cluster Construction.}
\label{sec:sparse_neuron}
Not all neurons in the overcomplete latent space are useful. Many are rarely activated across images or respond to visually inconsistent patterns. We therefore filter unreliable neurons and group the remaining ones into neuron clusters. We run the frozen VLM with the trained SAE over the training set and collect patch-level sparse activations $z_{i,p,j}$, where $j$ indexes the neuron. Neurons activated on fewer than $M_{\min}$ distinct images are discarded. For each retained neuron $j$, we aggregate its activations across patches within each image, select the top-$N$ images where neuron $j$ is most strongly activated, and localize the highest-activating patches by their spatial positions. The resulting patch regions $\mathcal{S}_j$ summarize the visual evidence associated with neuron $j$. We assign each neuron a short concept label by prompting an external model to summarize the shared visual pattern in $\mathcal{S}_j$ (see Appendix~\ref{app:prompt}), then encode labels into text embeddings and apply hierarchical clustering to group neurons into $K$ concept-level clusters $\{\mathcal{C}_k\}_{k=1}^{K}$. These labels are used only for cluster construction; after clustering, each cluster is represented only by its index and the corresponding set of neurons, forming the sparse neuron vocabulary used by NVF.

\subsection{Neuron-guided Visual Focus}
\label{sec:visual_focus}
Given the sparse neuron vocabulary constructed by SNS, NVF uses the language query to activate relevant visual concepts for the current image--question pair. It first predicts which neuron clusters are query-relevant, then localizes the patches where the selected concepts are most strongly expressed, and finally injects the activated evidence back into the corresponding vision tokens.
\begin{figure}[htbp]
    \centering
    \includegraphics[width=1\linewidth]{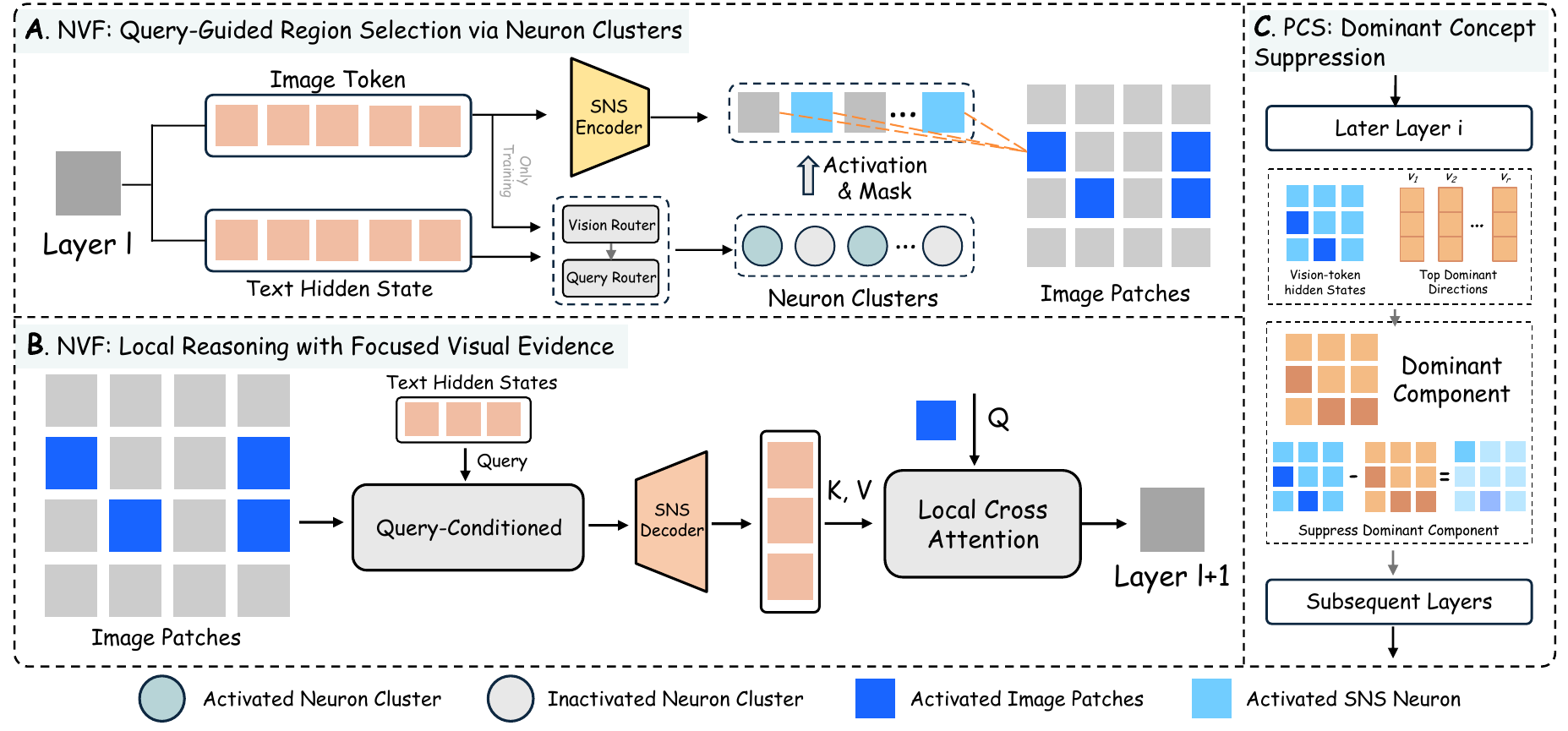}
    \caption{
    NVF and PCS. NVF selects query-relevant neuron cluster indices and active patches for local cross-attention, while PCS suppresses dominant perceptual components at a later layer.
    }
    \label{fig:nvf_pcs}
\end{figure}
\paragraph{Query-Guided Neuron Cluster Activation.}
Each neuron cluster is represented by an index $k \in \{1,\ldots,K\}$ and corresponds to a set of neurons. We obtain a query representation by pooling the textual hidden states and use a lightweight router $g_q(\cdot)$ to produce query-side cluster scores $\boldsymbol{\alpha}^{q}$ (Fig.~\ref{fig:nvf_pcs}A). To reduce reliance on language-only priors, a vision-side scorer $g_v(\cdot)$ produces image-grounded cluster scores $\boldsymbol{\alpha}^{v}$ from visual hidden states:

\begin{equation}
\boldsymbol{\alpha}^{q}
=
\sigma(g_q(\mathrm{Pool}(\mathbf{H}_{\mathrm{text}})))
\in [0,1]^K,
\qquad
\boldsymbol{\alpha}^{v}
=
\sigma(g_v(\mathrm{Pool}(\mathbf{H}_{\mathrm{vision}})))
\in [0,1]^K .
\end{equation}
At inference time, NVF selects the top-$K_{\mathrm{sel}}$ clusters according to $\boldsymbol{\alpha}^{q}$. During training, we obtain supervision labels $\mathbf{y}\in\{0,1\}^K$ by prompting an LLM annotator to identify which neuron clusters are relevant to each text query, based on the cluster descriptions constructed in Section~\ref{sec:sparse_neuron}. The router $g_q$ is then supervised with binary cross-entropy over $\mathbf{y}$, and a per-dimension symmetric divergence term aligns $\boldsymbol{\alpha}^{q}$ with $\boldsymbol{\alpha}^{v}$ to ensure that the selected clusters are grounded in image evidence:
\begin{equation}
\mathcal{L}_{\mathrm{align}}
=
\frac{1}{2}
\sum_{k=1}^{K}
\left[
\alpha_k^v
\log
\frac{\alpha_k^v}{\alpha_k^q}
+
\alpha_k^q
\log
\frac{\alpha_k^q}{\alpha_k^v}
\right].
\end{equation}
The final routing objective is $\mathcal{L}_{\mathrm{router}}=\mathcal{L}_{\mathrm{BCE}}(\boldsymbol{\alpha}^{q},\mathbf{y})+\lambda_{\mathrm{align}}\mathcal{L}_{\mathrm{align}}$, which encourages selected cluster indices to be both query-relevant and visually supported.

\paragraph{Local Reasoning with Focused Visual Evidence.}
For each selected cluster $c$, let $\mathcal{F}_c$ denote its set of neuron indices. NVF measures how strongly cluster $c$ is expressed at each patch token by summing the activations of its neurons: $a_{p}^{(c)}=\sum_{j \in \mathcal{F}_c} z_{p,j}$, where $z_{p,j}$ is the activation of neuron $j$ at patch token $p$. The top-$n$ patches with the highest $a_{p}^{(c)}$ are selected as active visual evidence for cluster $c$ (Fig.~\ref{fig:nvf_pcs}B). On these patches, we construct a cluster-specific sparse code $\mathbf{z}_{\mathrm{cluster}}$ by retaining only the neurons in $\mathcal{F}_c$ and masking all others. A query-conditioned refinement module then predicts a gated residual update:
\begin{equation}
\Delta \mathbf{z}
=
\beta_{\mathrm{ref}} \cdot
\mathbf{W}_{\Delta}
\left(
\sigma(\mathbf{W}_{g}\mathbf{z}_{\mathrm{cluster}})
\odot
\mathrm{MLP}(\mathbf{q}_{\mathrm{last}})
\right),
\end{equation}
where $\mathbf{q}_{\mathrm{last}}$ is the hidden state of the last textual token at layer $\ell$, and $\beta_{\mathrm{ref}}$ is a learnable scalar controlling the refinement magnitude. The refined sparse code is $\widetilde{\mathbf{z}}_{\mathrm{cluster}}=\mathbf{z}_{\mathrm{cluster}}+\Delta\mathbf{z}$. Each refined code is decoded back to the dense space via the frozen SAE decoder, weighted by its query-side score $\alpha_c^q$, and summed across selected clusters into $\mathbf{R}_{\mathrm{active}}$.

Finally, NVF injects $\mathbf{R}_{\mathrm{active}}$ into the selected active tokens through localized cross-attention:
\begin{equation}
\mathbf{H}_{\mathrm{active}}'
=
\mathbf{H}_{\mathrm{active}}
+
\gamma_{\mathrm{inj}} \cdot
\mathrm{LN}
\left(
\mathbf{W}_{o}
\mathrm{softmax}
\left(
\frac{\mathbf{Q}_{\mathrm{att}}\mathbf{K}_{\mathrm{att}}^{\top}}{\sqrt{d}}
\right)
\mathbf{V}_{\mathrm{att}}
\right),
\end{equation}
where $\mathbf{Q}_{\mathrm{att}}=\mathbf{W}_{q}\mathbf{H}_{\mathrm{active}}$, $\mathbf{K}_{\mathrm{att}}=\mathbf{W}_{k}\mathbf{R}_{\mathrm{active}}$, $\mathbf{V}_{\mathrm{att}}=\mathbf{W}_{v}\mathbf{R}_{\mathrm{active}}$, and $\gamma_{\mathrm{inj}}$ is a learnable injection scale. Only selected tokens are updated; all other vision tokens remain unchanged.

\subsection{Perceptual Concept Suppression}
\label{sec:pcs}

PCS addresses a side effect of focused activation. While NVF enhances query-relevant concepts, it may also concentrate vision-token representations along dominant directions, weakening cues such as spatial relations or fine-grained attributes. PCS mitigates this effect at a later layer $\ell' > \ell$ by estimating and attenuating these directions from vision-token geometry (Fig.~\ref{fig:nvf_pcs}C).

Let $\mathbf{H} \in \mathbb{R}^{N \times d}$ denote the vision-token hidden states at layer $\ell'$. We mean-center the tokens as $\bar{\mathbf{H}}=\mathbf{H}-\mathbf{1}\boldsymbol{\mu}^{\top}$, where $\boldsymbol{\mu}=\frac{1}{N}\sum_{i=1}^{N}\mathbf{H}_i \in \mathbb{R}^{d}$, to isolate directional structure from the global mean. We then compute a low-rank decomposition $\bar{\mathbf{H}}\approx\mathbf{U}\mathbf{\Sigma}\mathbf{V}^{\top}$ and take the top-$r$ right singular vectors $\mathbf{V}_r\in\mathbb{R}^{d\times r}$ as the dominant perceptual directions. The projected dominant component is $\mathbf{P}=\bar{\mathbf{H}}\mathbf{V}_r\mathbf{V}_r^{\top}$, and PCS suppresses it by:
\begin{equation}
\mathbf{H}'=\mathbf{H}-\eta_{\mathrm{pcs}}\,\mathbf{P},
\end{equation}
where $\eta_{\mathrm{pcs}}=\mathrm{softplus}(\eta_{\mathrm{param}})$ is a learnable non-negative scalar initialized near zero, allowing the model to learn the appropriate suppression strength during training. PCS complements NVF: NVF activates query-relevant concepts at layer $\ell$, while PCS preserves representational diversity at layer $\ell'$ by attenuating dominant directions that may suppress weaker visual cues.

\section{Experiments}

\subsection{Dataset}
\label{sec:dataset}
\paragraph{Training data.}
All training data are drawn exclusively from VQAv2~\cite{antol2015vqa} and are disjoint from the evaluation benchmarks. The SAE for SNS is trained on the full VQAv2 training split. For lightweight NVF modules and the PCS scalar, we construct three independent supervision sets, each with 5K image--question pairs sampled without replacement from VQAv2 training split to together with cluster-index labels. These labels are obtained by prompting an LLM annotator to identify query-relevant neuron clusters (Appendix~\ref{app:prompt}). For each backbone, we report the mean and standard deviation across the three resulting NeuronEye models.

\paragraph{Evaluation benchmarks.}
We evaluate on four vision-centric reasoning benchmarks covering spatial understanding, counting, relational perception, localization, and real-world visual reasoning. CV-Bench~\cite{tong2024cambrian1} serves as the primary benchmark, comprising Count, Depth, Distance, and Relation sub-tasks that directly assess structured visual reasoning. BLINK~\cite{fu2024blink} evaluates multi-view reasoning and spatial localization.  RealWorldQA~\cite{xai2024realworldqa} and MMStar~\cite{chen2024mmstar} provide complementary coverage of general visual question answering. No evaluation data are used during NeuronEye training.

\subsection{Implementation Details}
\label{sec:implementation}
We use Qwen2.5-VL-7B and LLaVA-1.6-7B as the base models, which are frozen throughout. NeuronEye constructs SNS at layer $\ell=8$ with $K=64$ neuron clusters, applies NVF at the same layer, and applies PCS at a later layer $\ell'=20$. After SNS construction, neuron-cluster assignments are kept fixed, and only the lightweight NVF and PCS scalar modules are updated. The NVF training procedure and hyperparameters are provided (Appendix~\ref{app:training_hyperparams}). We use greedy decoding for all generation-based evaluations. Experiments are conducted on NVIDIA RTX PRO 6000 GPUs.

\subsection{Benchmark Results across VLM Backbones}
\label{sec:main_results}

Table~\ref{tab:main_results} reports results across CV-Bench, BLINK, RealWorldQA, and MMStar. The upper block lists representative VLM baselines for reference; the lower block applies NeuronEye to Qwen2.5-VL-7B and LLaVA-1.6-7B to assess backbone portability. On Qwen2.5-VL-7B, NeuronEye improves CV-Bench overall by +3.1 and BLINK overall by +0.7, with the strongest gains on \emph{Distance} (+9.5), \emph{Multi-view} (+8.3), and \emph{Localization} (+3.3). On LLaVA-1.6-7B, improvements follow a similar pattern on spatial and relational sub-tasks, including \emph{Distance} (+6.9), \emph{Relation} (+12.1), \emph{Multi-view} (+7.7), and \emph{Localization} (+8.1), though performance decreases on \emph{Count} (-8.3) and \emph{Depth} (-3.3). This mixed pattern likely reflects backbone architectural differences in visual tokenization and patch granularity, which affect how patch-level sparse activations map to localized visual evidence. Across both backbones, the gains are consistently strongest on spatial, relational, and localization-sensitive tasks, aligning with the design of NeuronEye as a concept-level selective activation mechanism.

\begin{table*}[htbp]
\caption{Benchmark results across VLM backbones. The upper block lists representative VLMs for reference. The lower block shows NeuronEye applied to LLaVA-1.6-7B and Qwen2.5-VL-7B, with $\Delta$ denoting the improvement over each base model.}
\label{tab:main_results}
\centering
\small
\setlength{\tabcolsep}{4.5pt}
\renewcommand{\arraystretch}{1.08}
\resizebox{\textwidth}{!}{%
\begin{tabular}{lcccccccccc}
\toprule
& \multicolumn{5}{c}{CV-Bench}
& \multicolumn{3}{c}{BLINK}
& \multicolumn{2}{c}{Other Benchmarks} \\
\cmidrule(lr){2-6}
\cmidrule(lr){7-9}
\cmidrule(lr){10-11}
Model
& Overall & Count & Depth & Dist. & Rel.
& Overall & MV. & Loc.
& RealWorldQA & MMStar \\

\midrule
\multicolumn{11}{l}{\textit{Representative VLM backbones}} \\
DeepSeek-VL1~\cite{lu2024deepseek}
& 61.6 & 59.0 & 63.2 & 58.2 & 68.5
& 38.1 & 50.4 & 37.7
& 50.5 & 38.9 \\
Idefics3-8B-Llama3~\cite{laurencon2024building}
& 67.7 & 60.5 & 72.8 & 67.2 & 73.4
& 42.7 & 45.9 & 50.8
& 62.0 & 49.3 \\
Phi-4 Multimodal~\cite{abouelenin2025phi4mini}
& 71.7 & 68.5 & 74.2 & 70.8 & 75.1
& 49.7 & 48.1 & 56.6
& 61.8 & 59.7 \\
InternVL3-8B~\cite{zhu2025internvl3}
& 81.3 & 70.9 & 84.8 & 83.1 & 89.7
& 51.3 & 51.1 & 58.2
& 68.2 & 59.4 \\
Llava-OneVision~\cite{li2024llavaonevision}
& 76.0 & 67.3 & 80.3 & 78.5 & 80.8
& 46.2 & 57.1 & 54.9
& 66.7 & 68.3 \\

\midrule
\multicolumn{11}{l}{\textit{NeuronEye applied to different backbones}} \\
LLaVA-1.6-7B~\cite{liu2024llavanext}
& 64.3 & 63.3 & 77.8 & 54.5 & 63.3
& 36.6 & 40.7 & 38.0
& 61.6 & 36.9 \\
\textbf{+NeuronEye}
& 65.7 {\scriptsize$\pm$0.5} 
& 55.0 {\scriptsize$\pm$0.2} 
& 74.5 {\scriptsize$\pm$1.0}
& 61.4 {\scriptsize$\pm$0.9} 
& 75.4 {\scriptsize$\pm$0.5}
& 37.2 {\scriptsize$\pm$0.7} 
& 48.4 {\scriptsize$\pm$1.9} 
& 46.1 {\scriptsize$\pm$0.6}
& 60.7 {\scriptsize$\pm$0.7} 
& 36.9 {\scriptsize$\pm$0.1}\\
$\Delta$
& +1.4 & -8.3 & -3.3 & +6.9 & +12.1
& +0.6 & +7.7 & +8.1
& -0.9 & +0.0 \\
\midrule
Qwen2.5-VL-7B~\cite{bai2025qwen25vl}
& 78.5
& 67.1
& 87.2
& 76.2
& 87.2
& 52.1
& 55.6
& 53.3
& 68.5
& 58.8 \\
\textbf{+NeuronEye}
& 81.6 {\scriptsize$\pm$0.3} 
& 67.9 {\scriptsize$\pm$0.6}
& 87.0 {\scriptsize$\pm$0.4}
& 85.7 {\scriptsize$\pm$0.8}
& 89.5 {\scriptsize$\pm$0.5}
& 52.8 {\scriptsize$\pm$0.2}
& 63.9 {\scriptsize$\pm$0.4}
& 56.6 {\scriptsize$\pm$0.9}
& 68.8 {\scriptsize$\pm$0.4}
& 59.3 {\scriptsize$\pm$0.5}\\
$\Delta$
& \textbf{\textcolor{green!50!black}{+3.1}} 
& +0.8 
& -0.2 
& \textbf{\textcolor{green!50!black}{+9.5}} 
& +2.3
& +0.7 
& \textbf{\textcolor{green!50!black}{+8.3}} 
& \textbf{\textcolor{green!50!black}{+3.3}}
& +0.3 
& +0.5 \\
\bottomrule
\end{tabular}%
}

\end{table*}

\subsection{Comparison with Visual Representation Modulation Methods}
Table~\ref{tab:method_comparison} compares NeuronEye with vision-token reduction and representation-level methods on Qwen2.5-VL-7B under the same evaluation protocol. NeuronEye achieves the best overall performance on both CV-Bench and BLINK, with especially large gains on CV-Bench \emph{Distance} (+9.5) and BLINK \emph{Multi-view} (+8.3). Token reduction methods show consistent degradation on structure-sensitive sub-tasks, likely because discarding or merging patches eliminates visual evidence that cannot be recovered downstream. Representation-level methods maintain near-baseline performance but offer limited improvement, suggesting that uniform dense-state shifts lack the specificity needed for spatially demanding queries. NeuronEye's gains are concentrated precisely on these structure-sensitive tasks, supporting the hypothesis that concept-level selective activation provides finer control over which visual evidence is enhanced for a given query.
\vspace{-1em}
\begin{table}[htbp]
\caption{Comparison with visual representation modulation methods on Qwen2.5-VL-7B. All methods are evaluated under the same protocol.}
\label{tab:method_comparison}
\centering
\scriptsize
\setlength{\tabcolsep}{2.3pt}
\resizebox{\linewidth}{!}{%
\begin{tabular}{lcccccccccc}
\toprule
& \multicolumn{5}{c}{CV-Bench}
& \multicolumn{3}{c}{BLINK}
& \multicolumn{2}{c}{Other Benchmarks} \\
\cmidrule(lr){2-6}
\cmidrule(lr){7-9}
\cmidrule(lr){10-11}
Model
& Overall & Count & Depth & Dist. & Rel.
& Overall & MV. & Loc.
& RealWorldQA & MMStar \\
\midrule

\multicolumn{11}{l}{\textit{Vision-token reduction methods}} \\
FastV~\cite{chen2024fastv}
& 75.9 & 64.0 & 83.5 & 74.5 & 85.5
& 48.4 & 55.6 & 49.2
& 68.8 & 55.1 \\
SparseVLM~\cite{zhang2025sparsevlm}
& 69.8 & 54.3 & 74.7 & 68.8 & 86.3
& 46.9 & 55.6 & 59.0
& 50.9 & 53.8 \\
PruMerge~\cite{shang2025prumerge}
& 71.6 & 55.2 & 78.0 & 74.0 & 83.2
& 46.8 & 54.9 & 51.6
& 64.3 & 50.6 \\
MustDrop~\cite{liu2024mustdrop}
& 73.7 & 58.6 & 81.5 & 74.8 & 83.9
& 47.2 & 55.6 & 52.5
& 65.1 & 53.0 \\
PyramidDrop~\cite{xing2025conical}
& 72.5 & 60.2 & 78.7 & 70.8 & 84.2
& 49.1 & 55.6 & 52.5
& 62.2 & 50.3 \\
\midrule

\multicolumn{11}{l}{\textit{Representation-level intervention methods}} \\
SSL~\cite{hua2025ssl}
& 77.6 & 66.2 & 84.5 & 76.0 & 87.9
& 51.3 & 55.6 & 52.5
& 69.5 & 58.1 \\
SAVE~\cite{park2026save}
& 77.9 & 66.2 & 85.7 & 76.5 & 87.2
& 52.3 & 55.6 & 54.9
& 69.0 & 59.2 \\
VISTA~\cite{li2025vista}
& 78.0 & 66.0 & 85.8 & 76.2 & 88.0
& 51.7 & 55.6 & 53.3
& 69.4 & 59.6 \\

\midrule
\textbf{NeuronEye}
& \textbf{81.6} {\tiny$\pm$0.3} 
& \textbf{67.9} {\tiny$\pm$0.6}
& \textbf{87.0} {\tiny$\pm$0.4}
& \textbf{85.7} {\tiny$\pm$0.8}
& \textbf{89.5} {\tiny$\pm$0.5}
& \textbf{52.8} {\tiny$\pm$0.2}
& \textbf{63.9} {\tiny$\pm$0.4}
& 56.6 {\tiny$\pm$0.9}
& 68.8 {\tiny$\pm$0.4}
& 59.3 {\tiny$\pm$0.5}\\

\bottomrule
\end{tabular}%
}

\end{table}

\subsection{Visualization of Sparse Neurons and Visual Focus}
We visualize sparse neuron selectivity and query-guided visual focus to examine how NeuronEye localizes concept-level evidence.

\begin{figure}[htbp]
\centering
\begin{subfigure}[b]{0.32\linewidth}
    \includegraphics[width=\linewidth]{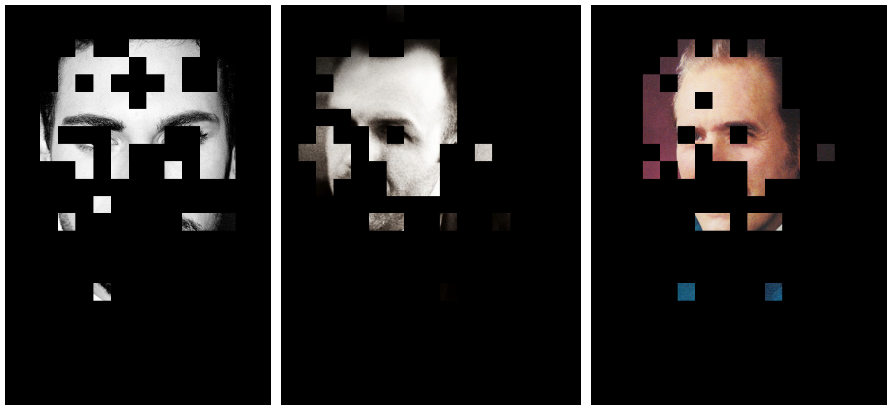}
    \caption{Human-face feature}
\end{subfigure}
\hfill
\begin{subfigure}[b]{0.32\linewidth}
    \includegraphics[width=\linewidth]{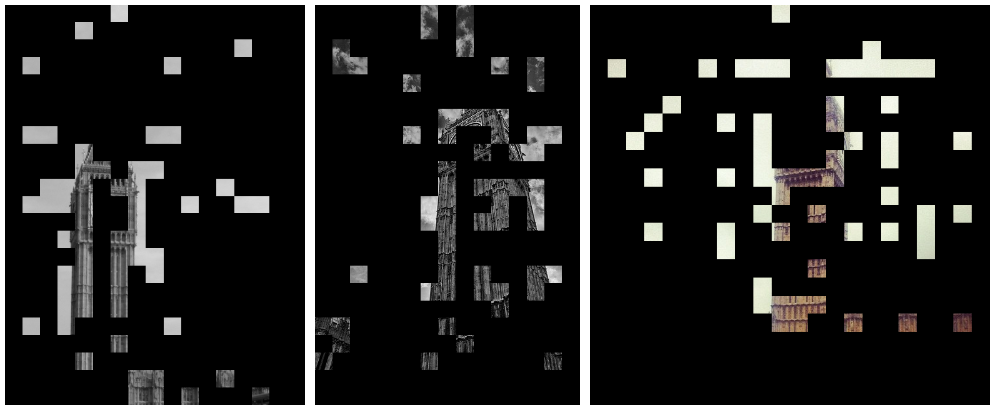}
    \caption{Tower feature}
\end{subfigure}
\hfill
\begin{subfigure}[b]{0.32\linewidth}
    \includegraphics[width=\linewidth]{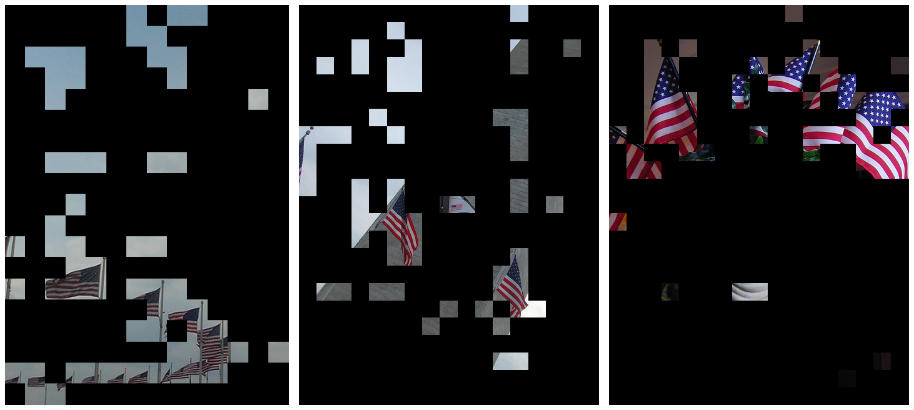}
    \caption{Flag feature}
\end{subfigure}
\caption{Neuron-level feature visualization. Each subfigure shows top-activating patches for one SAE neuron, with non-activating regions masked.}
\label{fig:neuron_vis}
\end{figure}

\textbf{Neuron-level features.}
Fig.~\ref{fig:neuron_vis} shows that individual SNS neurons activate consistently on semantically coherent regions, such as faces, towers, and flags. These examples support using SNS neurons as a sparse visual vocabulary for concept-level organization. Additional neuron visualizations are provided in the appendix (Appendix~\ref{app:neuron_features}).

\begin{figure}[htbp]
    \centering
    \captionsetup[subfigure]{skip=2pt}
    \setlength{\abovecaptionskip}{3pt}
    \setlength{\belowcaptionskip}{0pt}

    \begin{subfigure}{1\linewidth}
        \centering
        \includegraphics[width=\linewidth]{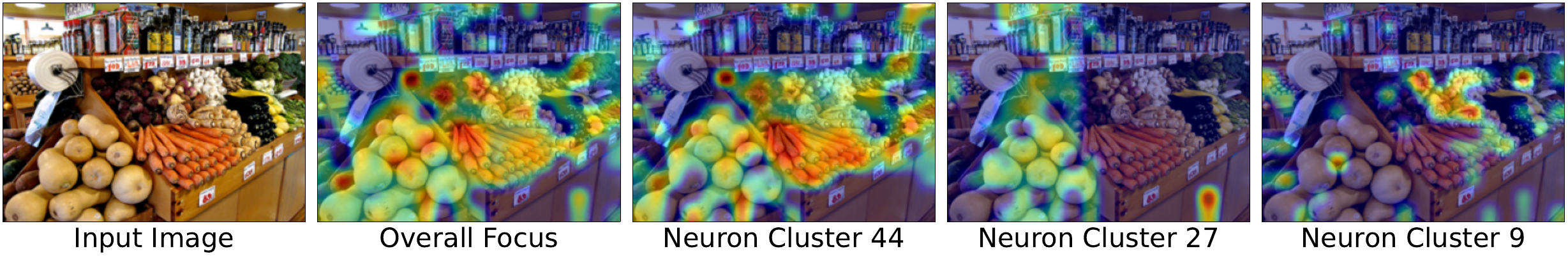}
        \caption{Is there organic food in this store?}
        \label{fig:focus_organic_food}
    \end{subfigure}

    \vspace{0.2em}

    \begin{subfigure}{0.48\linewidth}
        \centering
        \includegraphics[width=\linewidth]{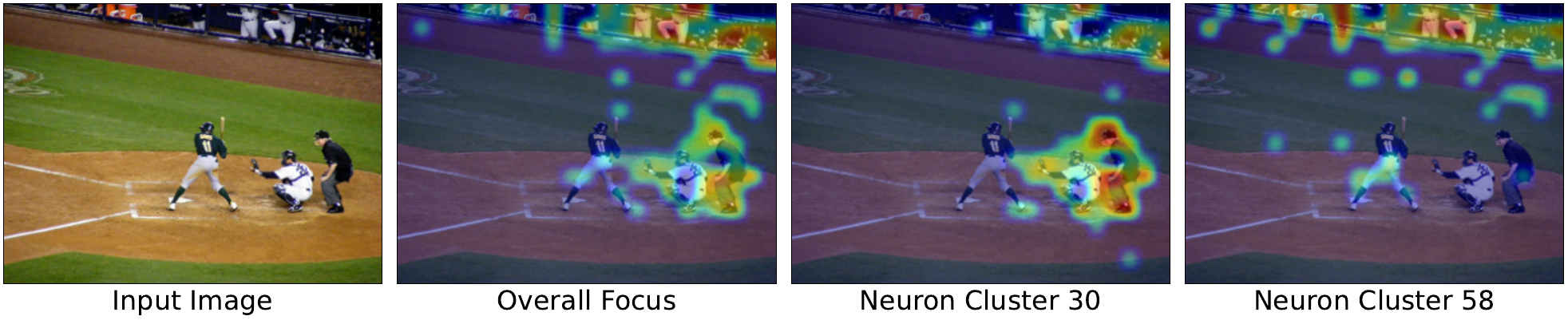}
        \caption{Are all the players wearing black shirts?}
        \label{fig:focus_black_shirts}
    \end{subfigure}
    \hfill
    \begin{subfigure}{0.48\linewidth}
        \centering
        \includegraphics[width=\linewidth]{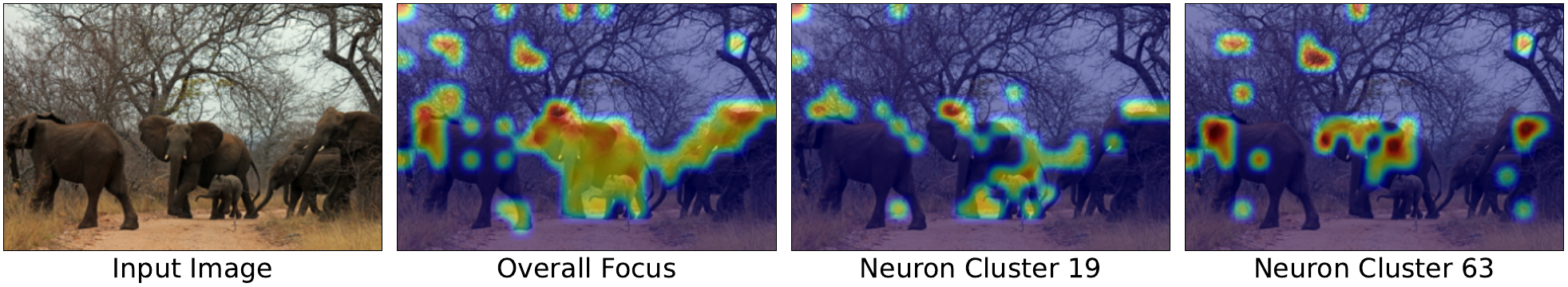}
        \caption{Is the small elephant touching the big elephant?}
        \label{fig:focus_elephant}
    \end{subfigure}

    \caption{Question-guided visual focus examples. Each shows the input image, overall focus map, and individual neuron cluster activations.}
    \label{fig:focus_combined}
\end{figure}








Fig.~\ref{fig:focus_combined} visualizes the spatial focus produced by NVF. Given a question, the router selects a small set of neuron clusters and actively triggers the corresponding neurons within these clusters. Their patch-level activations highlight localized image regions expressing the selected sparse features. For the query ``Is there organic food in this store?'', NVF focuses on shelves and packaged items; for ``Are all the players wearing black shirts?'', it shifts toward clothing and human-body regions; and for ``Is the small elephant touching the big elephant?'', it emphasizes the two elephants and their contact region. The resulting focus maps are sparse and localized, indicating that NVF operates through query-selected neuron clusters and patch-level activations rather than global image reweighting. Additional focus examples are provided in the appendix (Appendix~\ref{app:neuron_focus}).

\subsection{Ablation Studies}
\vspace{-1em}
\begin{table}[htbp]
\centering
\caption{Ablation study on component contributions.}
\small
\setlength{\tabcolsep}{2pt}
\begin{tabular}{lcccccccc}
\toprule
& \multicolumn{5}{c}{CV-Bench}
& \multicolumn{3}{c}{BLINK} \\
\cmidrule(lr){2-6}
\cmidrule(lr){7-9}
Setting
& Overall & Count & Depth & Dist. & Rel.
& Overall & MV. & Loc. \\
\midrule
Baseline (Qwen2.5-VL-7B)
& 78.5 & 67.1 & 87.2 & 76.2 & 87.2
& 52.1 & 55.6 & 53.3 \\
Baseline + FT
& 76.5 {\tiny$\pm$0.3}& 66.6 {\tiny$\pm$0.2}& 86.7 {\tiny$\pm$0.3}& 69.8 {\tiny$\pm$0.9}& 86.5{\tiny$\pm$0.6}
& 51.1 {\tiny$\pm$0.1}& 55.6 {\tiny$\pm$1.9}& 55.7 {\tiny$\pm$0.4}\\

+ Dense Cross-Attn + PCS
& 79.4 {\tiny$\pm$0.3}& 66.2 {\tiny$\pm$0.6}& 87.0 {\tiny$\pm$0.4}& 79.5 {\tiny$\pm$1.2}& 88.8{\tiny$\pm$0.3}
& 51.3 {\tiny$\pm$0.3}& 55.6 {\tiny$\pm$1.9}& 54.9 {\tiny$\pm$1.8}\\

+ Random SNS + NVF
& 78.6 {\tiny$\pm$0.4}& 67.6 {\tiny$\pm$0.5}& 86.5 {\tiny$\pm$0.6}& 76.6 {\tiny$\pm$1.0}& 86.4{\tiny$\pm$0.6}
& 51.6 {\tiny$\pm$0.3}& 55.6 {\tiny$\pm$1.5}& 54.9 {\tiny$\pm$1.0}\\

+ PCS only
& 78.6 {\tiny$\pm$0.4} & 67.3 {\tiny$\pm$0.4}& 86.8 {\tiny$\pm$0.3}& 76.8 {\tiny$\pm$1.4}& 87.2 {\tiny$\pm$0.6}
& 51.3 {\tiny$\pm$0.2}& 55.6 {\tiny$\pm$1.9}& 54.9 {\tiny$\pm$0.4}\\

+ SNS + NVF
& 80.9 {\tiny$\pm$0.1} & 66.8 {\tiny$\pm$0.6}& 87.2 {\tiny$\pm$0.5}& 84.0 {\tiny$\pm$0.4}& 89.5 {\tiny$\pm$0.5}
& 53.0 {\tiny$\pm$0.3}& 55.6 {\tiny$\pm$2.2}& 55.7 {\tiny$\pm$3.0}\\

+ NeuronEye
& 81.6 {\tiny$\pm$0.3} 
& 67.9 {\tiny$\pm$0.6}
& 87.0 {\tiny$\pm$0.4}
& 85.7 {\tiny$\pm$0.8}
& 89.5 {\tiny$\pm$0.5}
& 52.8 {\tiny$\pm$0.2}
& 63.9 {\tiny$\pm$0.4}
& 56.6 {\tiny$\pm$0.9} \\
\bottomrule
\end{tabular}
\label{tab:ablation}
\end{table}

Table~\ref{tab:ablation} summarizes the contribution of each component. Direct fine-tuning of the lightweight modules without sparse intervention (Baseline + FT) does not improve performance and substantially reduces CV-Bench Distance, suggesting that limited VQA fine-tuning alone is insufficient to produce the observed spatial reasoning gains.

Replacing semantically organized neuron clusters with random assignments (Random SNS + NVF) yields near-baseline performance, indicating that the gains do not come merely from added parameters or reconstructed-feature injection. Rather, structured clusters are needed to route sparse visual evidence meaningfully. PCS alone produces only marginal changes, suggesting that attenuating dominant directions is insufficient without query-guided feature selection. SNS + NVF improves CV-Bench Overall and Distance, while the full NeuronEye configuration further improves Distance and BLINK Multi-view. Together, these results support the complementary design of NeuronEye: NVF activates localized query-relevant sparse features, while PCS later attenuates dominant directions to preserve representational diversity and weaker cues.

\begin{table}[htbp]
\centering
\caption{Effect of insertion layer. Layer 8 achieves the best results, suggesting a balance between low-level perceptual features and high-level semantic abstraction.}
\small
\setlength{\tabcolsep}{4pt}
\begin{tabular}{lcccccccc}
\toprule
& \multicolumn{5}{c}{CV-Bench}
& \multicolumn{3}{c}{BLINK} \\
\cmidrule(lr){2-6}
\cmidrule(lr){7-9}
Layer $\ell$
& Overall & Count & Depth & Dist. & Rel.
& Overall & MV. & Loc. \\
\midrule
4  
& 80.5 {\tiny$\pm$0.1}
& 67.3 {\tiny$\pm$0.2}
& 87.0 {\tiny$\pm$0.6}
& 83.0 {\tiny$\pm$0.4}
& 88.5 {\tiny$\pm$0.3}
& 51.6 {\tiny$\pm$0.3}
& 57.2 {\tiny$\pm$1.2}
& 56.6 {\tiny$\pm$1.1}\\

8  
& 81.6 {\tiny$\pm$0.3} 
& 67.9 {\tiny$\pm$0.6}
& 87.0 {\tiny$\pm$0.4}
& 85.7 {\tiny$\pm$0.8}
& 89.5 {\tiny$\pm$0.5}
& 52.8 {\tiny$\pm$0.2}
& 63.9 {\tiny$\pm$0.4}
& 56.6 {\tiny$\pm$0.9} \\

12 & 78.3 {\tiny$\pm$0.4} & 65.2 {\tiny$\pm$0.8}& 85.7 {\tiny$\pm$0.8}& 79.5 {\tiny$\pm$1.2}& 86.3 {\tiny$\pm$0.3}
& 51.3 {\tiny$\pm$0.2}& 55.6 {\tiny$\pm$0.1}& 54.1 {\tiny$\pm$0.1}\\

16 & 79.0 {\tiny$\pm$0.3}& 67.7 {\tiny$\pm$0.6} & 87.2 {\tiny$\pm$0.2} & 76.9 {\tiny$\pm$0.5} & 87.7 {\tiny$\pm$0.6}
& 51.8{\tiny$\pm$0.5} & 56.9{\tiny$\pm$1.9} & 56.3{\tiny$\pm$0.5} \\

20 & 78.6 {\tiny$\pm$0.4}& 66.8 {\tiny$\pm$0.3}& 86.8 {\tiny$\pm$0.4}& 77.0 {\tiny$\pm$1.1}& 86.9 {\tiny$\pm$0.6}
& 51.7 {\tiny$\pm$0.1}& 55.6 {\tiny$\pm$1.9}& 55.7 {\tiny$\pm$1.3}\\
\bottomrule
\end{tabular}
\label{tab:layer_ablation}
\end{table}

Table~\ref{tab:layer_ablation} demonstrates the effect of the NVF insertion layer. Performance peaks at layer 8, especially on CV-Bench Distance and BLINK Multi-view. Earlier insertion at layer 4 remains competitive but is slightly weaker, suggesting that the representation may not yet provide sufficient semantic abstraction for reliable cluster routing. Deeper layers from 12 to 20 degrade toward the baseline, likely because spatial details have become increasingly compressed. 

\vspace{-1em}

\begin{table}[htbp]
\centering
\caption{Effect of the number of neuron clusters K. Too few clusters conflate functionally distinct neurons, while too many fragment coherent functional groups and complicate routing.}
\small
\setlength{\tabcolsep}{4pt}
\begin{tabular}{lcccccccc}
\toprule
& \multicolumn{5}{c}{CV-Bench}
& \multicolumn{3}{c}{BLINK} \\
\cmidrule(lr){2-6}
\cmidrule(lr){7-9}
Clusters
& Overall & Count & Depth & Dist. & Rel.
& Overall & MV. & Loc. \\
\midrule
32  
& 80.4 {\tiny$\pm$0.2}
& 68.4 {\tiny$\pm$0.5}
& 86.1 {\tiny$\pm$0.4}
& 82.1 {\tiny$\pm$1.2}
& 88.2 {\tiny$\pm$0.5}
& 52.0 {\tiny$\pm$0.2}
& 57.2 {\tiny$\pm$1.5}
& 53.9 {\tiny$\pm$2.6}\\

64  
& 81.6 {\tiny$\pm$0.3} 
& 67.9 {\tiny$\pm$0.6}
& 87.0 {\tiny$\pm$0.4}
& 85.7 {\tiny$\pm$0.8}
& 89.5 {\tiny$\pm$0.5}
& 52.8 {\tiny$\pm$0.2}
& 63.9 {\tiny$\pm$0.4}
& 56.6 {\tiny$\pm$0.9} \\

128 
& 80.3 {\tiny$\pm$0.1}
& 66.4 {\tiny$\pm$1.1}
& 87.1 {\tiny$\pm$0.4}
& 82.8 {\tiny$\pm$0.7}
& 88.1 {\tiny$\pm$0.4}
& 51.7 {\tiny$\pm$0.3}
& 57.8 {\tiny$\pm$1.7}
& 58.3 {\tiny$\pm$1.3}\\

\bottomrule
\end{tabular}
\label{tab:cluster_ablation}
\end{table}

Table~\ref{tab:cluster_ablation} examines the effect of the number of neuron clusters. $K=64$ achieves the best overall trade-off, with the strongest results on CV-Bench Distance and BLINK Multi-view. 
With fewer clusters ($K=32$), neurons corresponding to different visual concepts may be grouped together, reducing routing specificity. 
With more clusters ($K=128$), neurons corresponding to the same or closely related concept may be split across multiple clusters, making cluster prediction less stable and reducing the coherence of concept-level activation.

\section{Discussion}

\paragraph{Limitations.}
NeuronEye improves tasks that rely on localized spatial or relational evidence, but its effectiveness depends on the coverage of the learned neuron vocabulary. Since the SAE and neuron clusters are trained on VQAv2 and then frozen, the sparse neuron vocabulary reflects general-domain visual concepts and may lack coverage of domain-specific ones. This limitation is evident on a medical domain VQA dataset (Appendix~\ref{app:medical_domain_shift}), OmniMedVQA-Mini~\cite{hu2024omnimedvqa}, where NeuronEye reduces overall accuracy from 65.30\% to 63.10\%, with larger drops on disease diagnosis (-2.80), lesion grading (-2.18), and other biological attributes (-5.51). NeuronEye also relies on LLM-generated cluster-index labels for routing supervision at scale. We manually verified approximately 20\% of the generated labels, but misalignment between query intent and selected neuron clusters may persist in the remainder, potentially degrading the model's performance. Additionally, NeuronEye introduces approximately 20\% inference latency overhead and 42\% peak memory increase relative to the base model, primarily due to frozen SAE encoding and localized cross-attention (Appendix~\ref{app:efficiency}).


\paragraph{Future directions.}
Future work could expand the sparse neuron space using larger and more diverse training corpora, or construct domain-specific neuron vocabularies for specialized settings such as medical imaging. Extending sparse routing across multiple layers may enable coordinated modulation across levels of visual abstraction. Learning neuron clusters directly from data could further reduce reliance on offline concept labeling and improve robustness under distribution shift.

\section{Conclusion}

We presented NeuronEye, a plug-in framework that decomposes dense VLM representations into a sparse, concept-level neuron vocabulary and selectively activates query-relevant visual concepts at inference, without modifying or retraining the VLM backbone. Experiments across four benchmarks and two backbones demonstrate consistent improvements on spatial, relational, and multi-view reasoning. Our results suggest that sparse autoencoder features can serve not only as interpretability tools but as active interfaces for modulating concept-level visual reasoning.

\bibliographystyle{unsrt}
\bibliography{references}

\clearpage
\appendix

\section{Offline Interpretation Prompts}
\label{app:prompt}

\begin{figure}[htbp]
    \centering
    \includegraphics[width=0.98\linewidth]{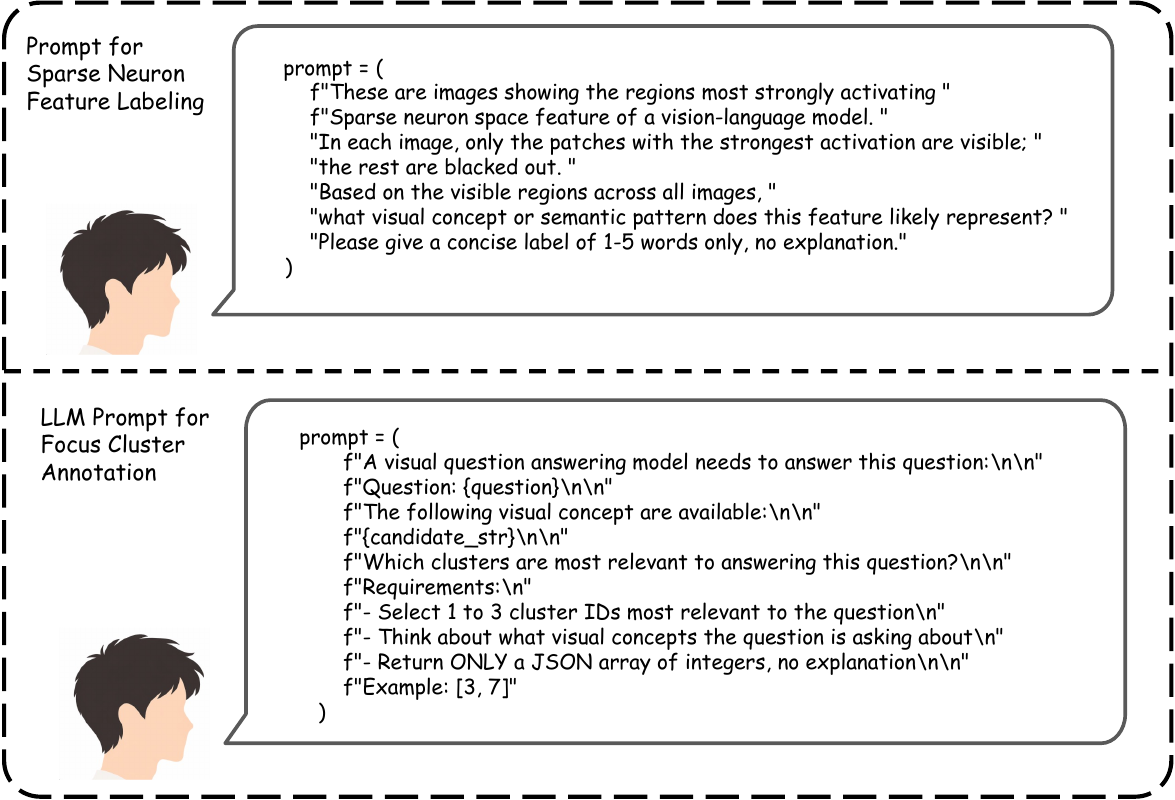}
    \caption{Offline prompts used for sparse neuron interpretation and cluster-index supervision.}
    \label{fig:label_prompt}
\end{figure}

Figure~\ref{fig:label_prompt} shows the offline prompts used for sparse-neuron interpretation and cluster-index supervision. We use Claude 3.7 Sonnet, a multimodal model with vision capabilities accessed through the Anthropic API, as an external annotator. The first prompt assigns short concept labels to sparse neurons based on representative image regions that strongly activate each neuron. The second prompt maps training questions to relevant cluster indices using the offline cluster inventory. The resulting annotations are used only before training to construct neuron clusters and generate cluster-index labels; they are not provided to the router, the VLM backbone, or any inference-time component.

\clearpage
\section{Training, Inference, and Hyperparameter Details}
\label{app:training_hyperparams}

Training follows a two-stage schedule. In Stage~1, the router and vision-side scorer are trained using frozen hidden states, optimizing the cluster-index prediction loss and the query--vision alignment loss before any feature injection is introduced. This stage stabilizes the query-to-cluster mapping. In Stage~2, the full NeuronEye pipeline is enabled: NVF performs sparse feature localization, query-conditioned refinement, and localized injection, while PCS suppresses dominant directions at the later layer. The lightweight components are optimized with the combined objective
\[
\mathcal{L}
=
\mathcal{L}_{\mathrm{LM}}
+
\lambda_{\mathrm{router}}\mathcal{L}_{\mathrm{router}} .
\]
Throughout both stages, the VLM backbone, pretrained SAE, and neuron-cluster assignments remain fixed. Only the lightweight NVF modules and the PCS scalar are updated.

At inference time, NeuronEye runs in a single modified forward pass. NVF selects neuron clusters, localizes and refines the corresponding sparse features, and injects localized updates at layer $\ell$, while PCS suppresses dominant directions at layer $\ell'$. No external annotator, retrieval module, parameter update, or multi-pass decoding is used during inference.

\begin{table}[htbp]
\centering
\caption{Hyperparameter settings for all experiments.}
\label{tab:hyperparams}
\begin{tabular}{lcc}
\toprule
\textbf{Component} & \textbf{Hyperparameter} & \textbf{Value} \\
\midrule
\multirow{4}{*}{Sparse Neuron Space} & Expansion ratio $\alpha$ & 32 \\
 & Sparsity level $k$ & 32 \\
 & Sparsity coefficient $\lambda_s$ & 0.05 \\
 & Min. image threshold $M_{\min}$ & 3 \\
\midrule
\multirow{5}{*}{Neuron-guided Visual Focus} & Top-$n$ patches per cluster & 60 \\
 & Top-$K_{\mathrm{sel}}$ clusters & 5 \\
 & $\gamma_{\mathrm{inj}}$ init & 0.75 \\
 & Bottleneck dim & 128 \\
 & $\beta_{\mathrm{ref}}$ init & $-3.0$ \\
\midrule
\multirow{2}{*}{PCS} & Suppressed directions $r$ & 2 \\
 & $\eta_{\mathrm{param}}$ init & $-3.0$ \\
\midrule
\multirow{5}{*}{Training} & Learning rate & $1\times10^{-4}$ \\
 & Gradient accumulation & 6 \\
 & Training samples & 5K \\
 & $\lambda_{\mathrm{router}}$ & 0.5 \\
 & $\lambda_{\mathrm{align}}$ & 0.3 \\
\bottomrule
\end{tabular}
\end{table}

\clearpage
\section{Additional Sparse Neuron Visualizations}
\label{app:neuron_features}

Figures in this section provide additional examples of individual SNS neurons and their top-activating patches. Each visualization shows image regions that strongly activate one sparse latent dimension, with non-activating regions masked. These examples illustrate that many retained neurons correspond to localized and semantically coherent visual patterns, supporting their organization into concept-level neuron clusters.

\begin{figure}[htbp]
\centering
\begin{subfigure}[b]{0.48\linewidth}
    \includegraphics[width=\linewidth]{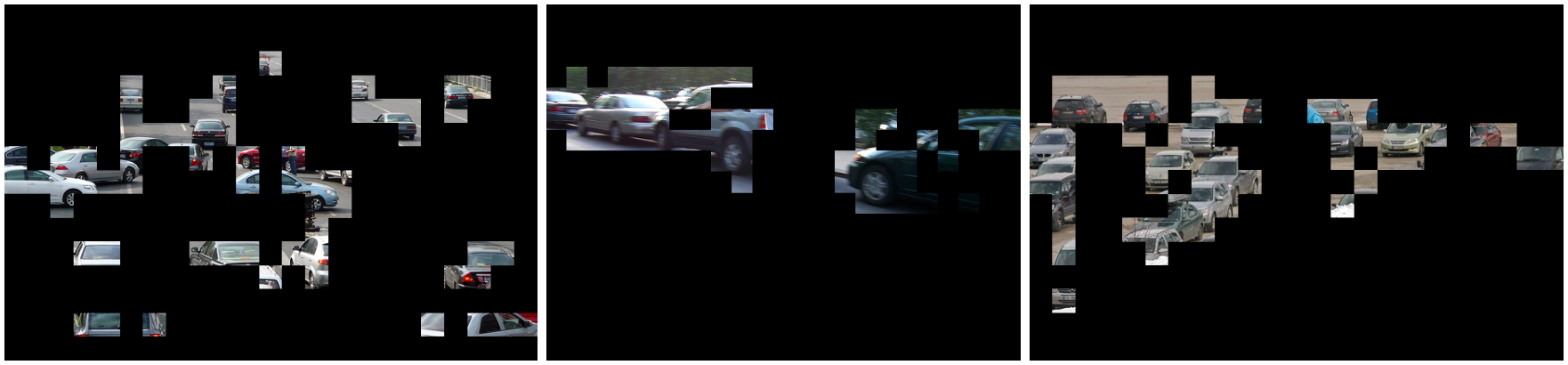}
    \caption{Vehicle feature}
\end{subfigure}
\hfill
\begin{subfigure}[b]{0.48\linewidth}
    \includegraphics[width=\linewidth]{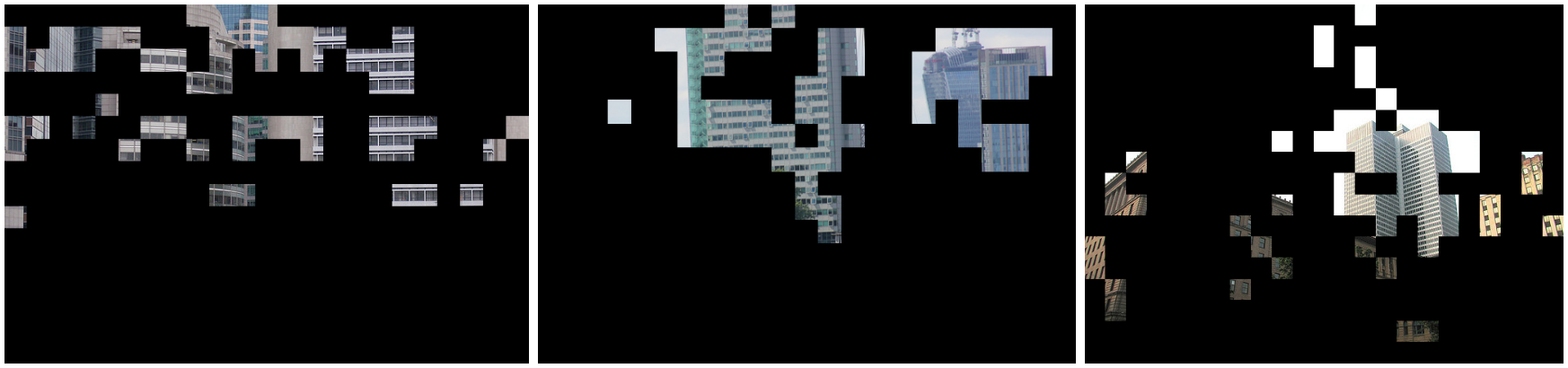}
    \caption{Building feature}
\end{subfigure}

\vspace{0.5em}
\begin{subfigure}[b]{0.48\linewidth}
    \includegraphics[width=\linewidth]{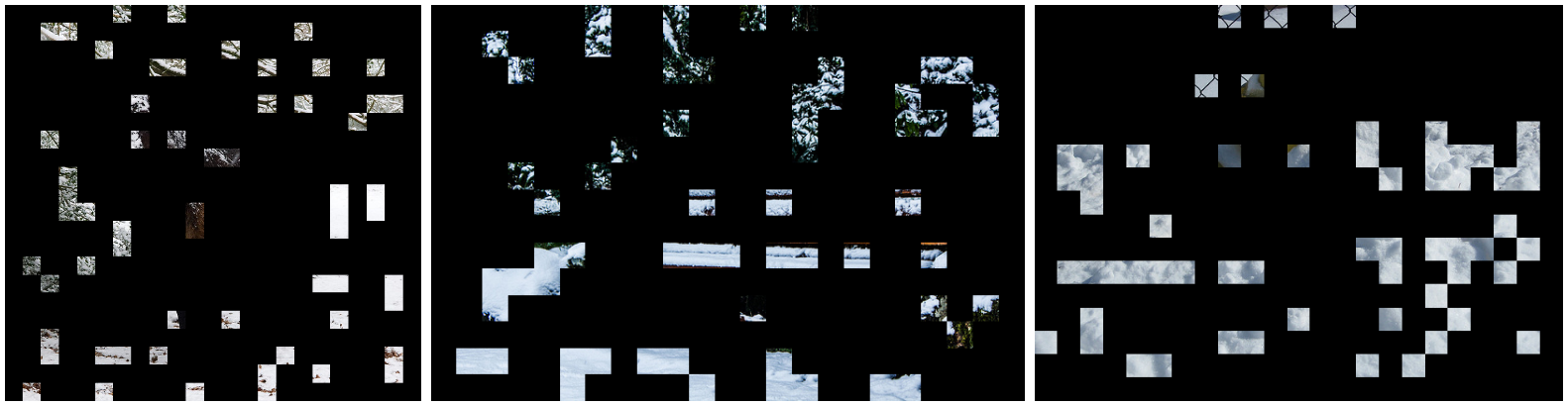}
    \caption{Snow feature}
\end{subfigure}
\hfill
\begin{subfigure}[b]{0.48\linewidth}
    \includegraphics[width=\linewidth]{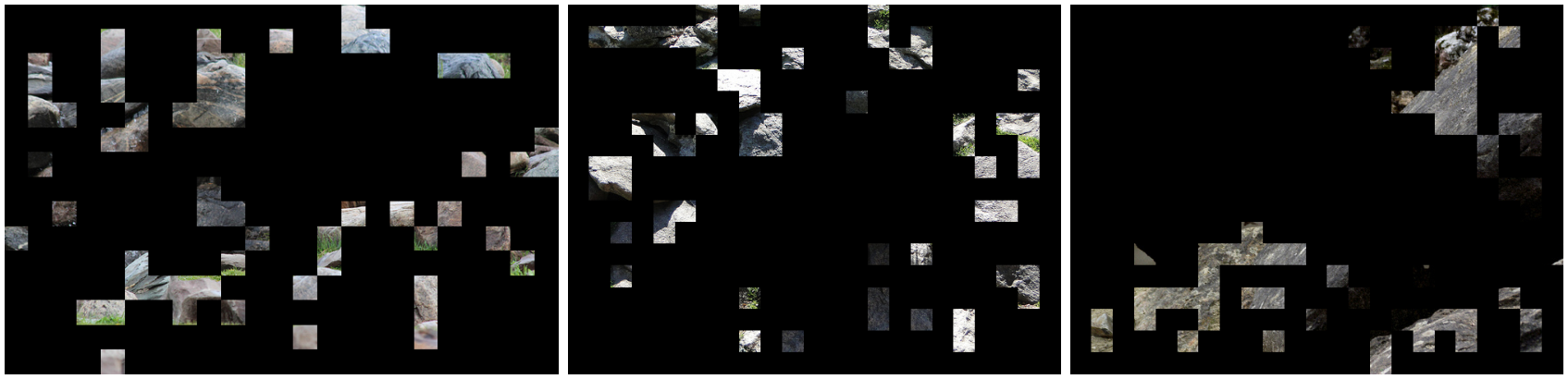}
    \caption{Rocky terrain feature}
\end{subfigure}

\vspace{0.5em}

\begin{subfigure}[b]{0.48\linewidth}
    \includegraphics[width=\linewidth]{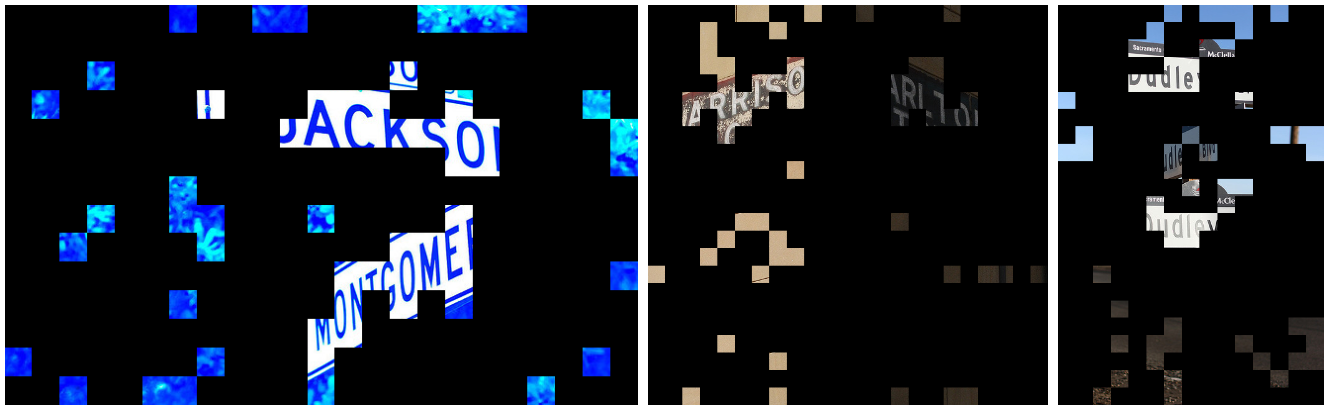}
    \caption{Street signs feature}
\end{subfigure}
\hfill
\begin{subfigure}[b]{0.48\linewidth}
    \includegraphics[width=\linewidth]{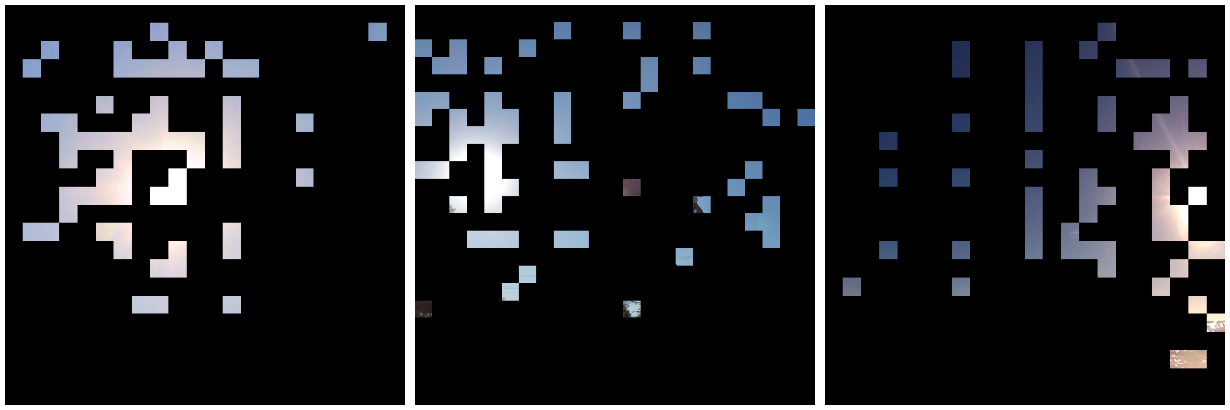}
    \caption{Sky and sun feature}
\end{subfigure}

\vspace{0.5em}

\begin{subfigure}[b]{0.48\linewidth}
    \includegraphics[width=\linewidth]{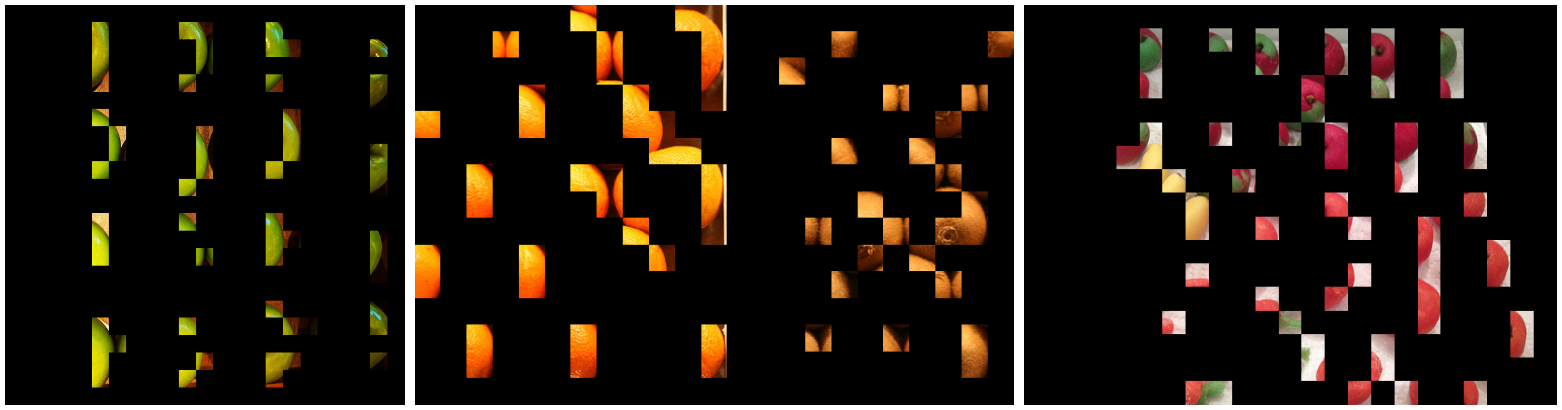}
    \caption{Fruit feature}
\end{subfigure}
\hfill
\begin{subfigure}[b]{0.48\linewidth}
    \includegraphics[width=\linewidth]{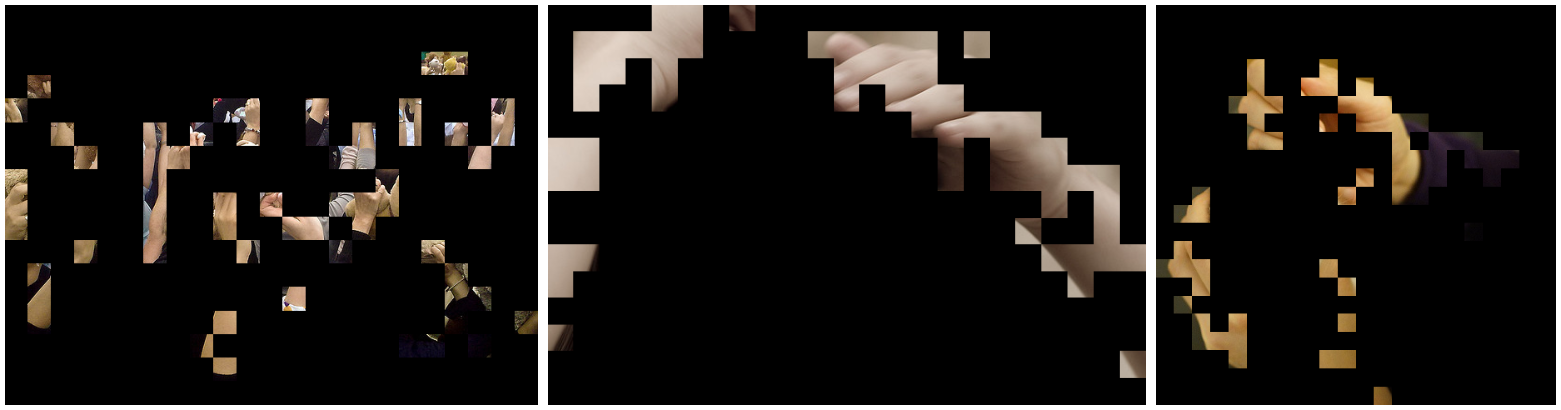}
    \caption{Hands feature}
\end{subfigure}

\vspace{0.5em}

\begin{subfigure}[b]{0.48\linewidth}
    \includegraphics[width=\linewidth]{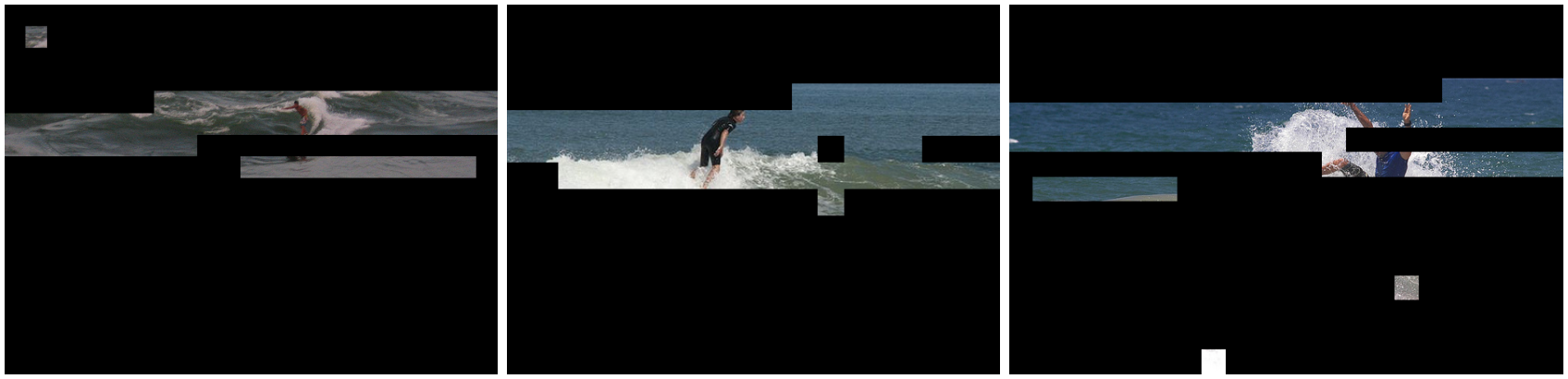}
    \caption{Water sports feature}
\end{subfigure}
\hfill
\begin{subfigure}[b]{0.48\linewidth}
    \includegraphics[width=\linewidth]{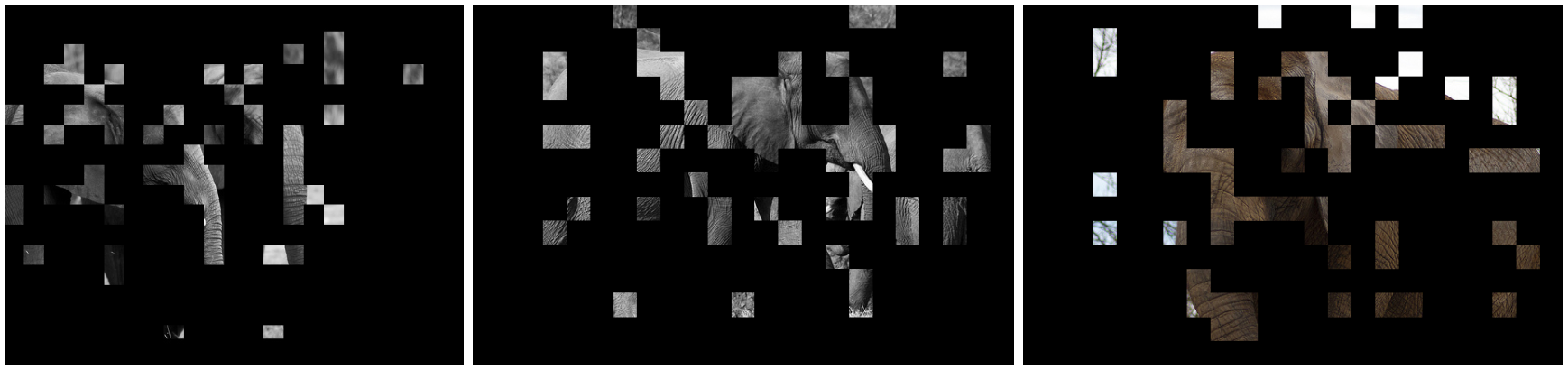}
    \caption{Elephant feature}
\end{subfigure}

\vspace{0.5em}

\begin{subfigure}[b]{0.48\linewidth}
    \includegraphics[width=\linewidth]{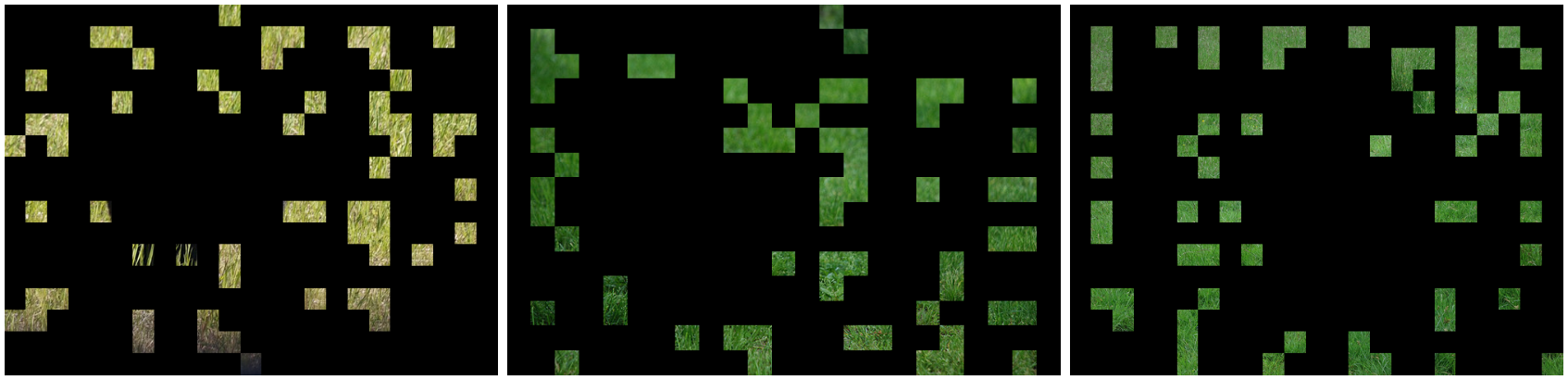}
    \caption{Grass feature}
\end{subfigure}
\hfill
\begin{subfigure}[b]{0.48\linewidth}
    \includegraphics[width=\linewidth]{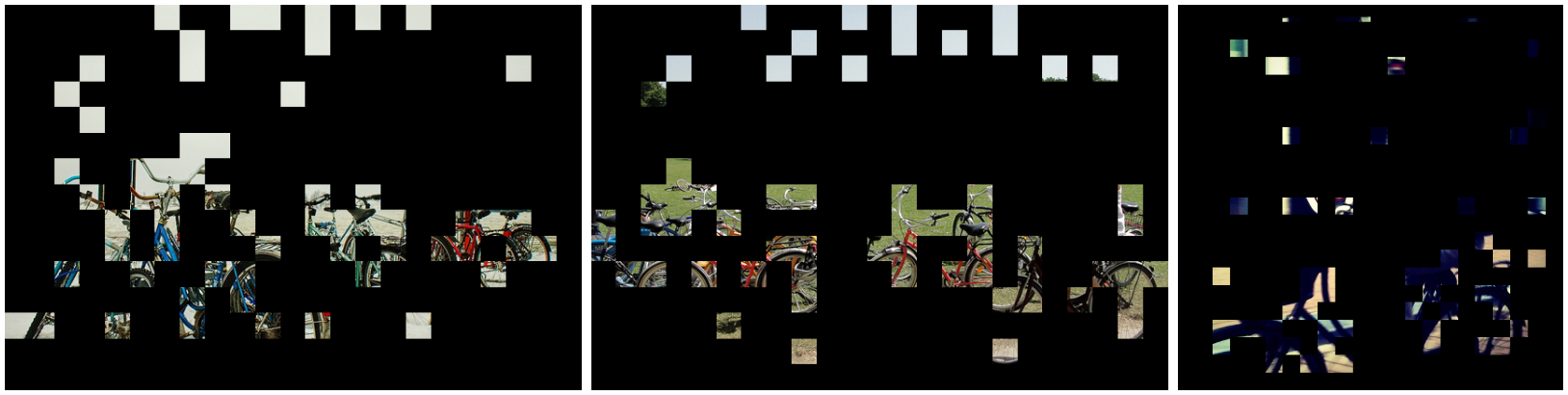}
    \caption{Bicycle feature}
\end{subfigure}

\vspace{0.5em}

\begin{subfigure}[b]{0.48\linewidth}
    \includegraphics[width=\linewidth]{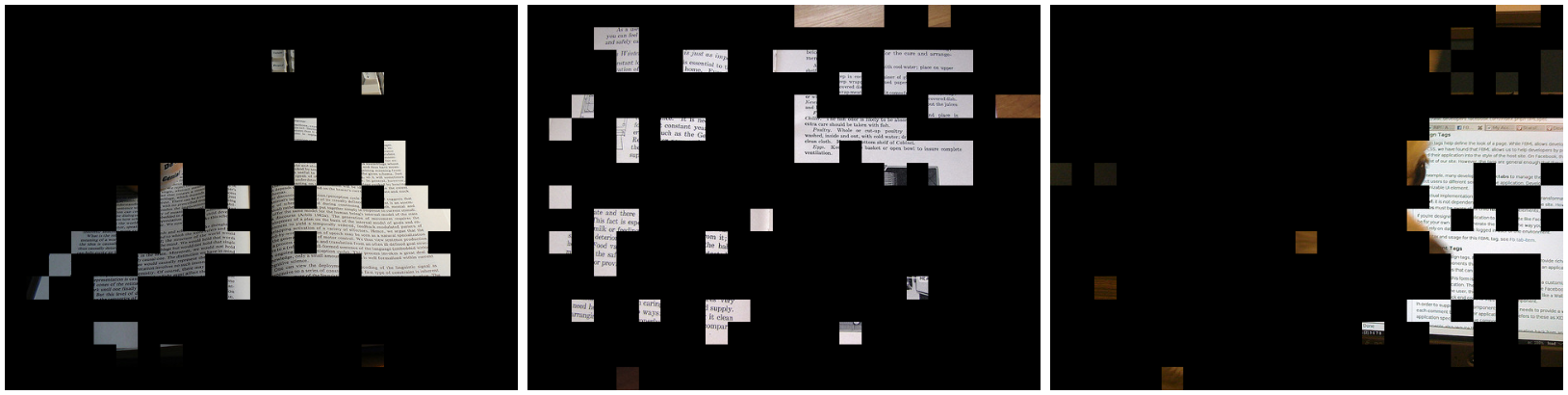}
    \caption{Textual information feature}
\end{subfigure}
\hfill
\begin{subfigure}[b]{0.48\linewidth}
    \includegraphics[width=\linewidth]{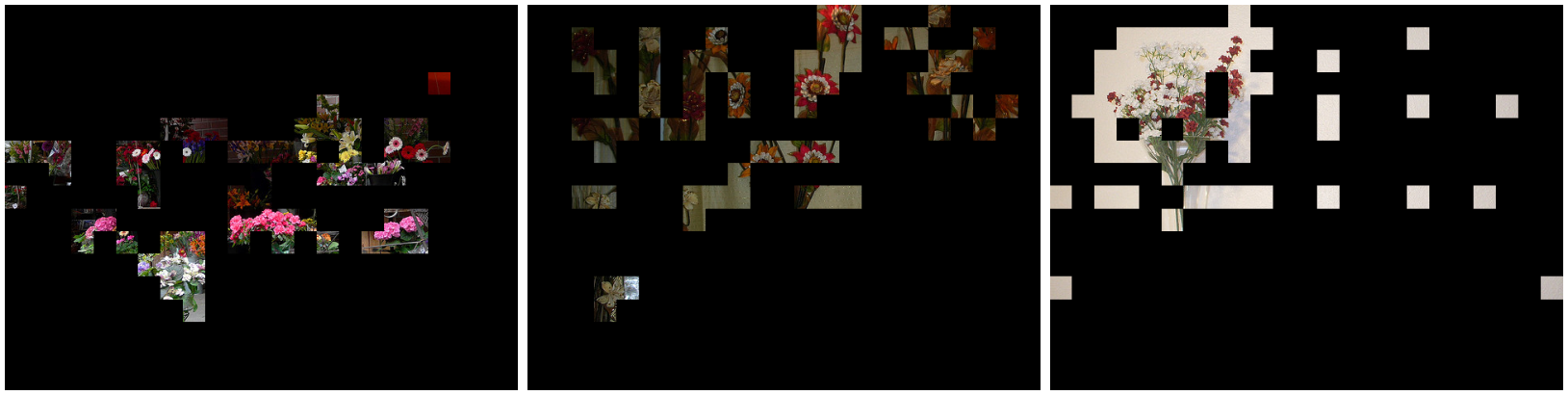}
    \caption{Flowers feature}
\end{subfigure}

\vspace{0.5em}

\caption{Additional sparse neuron visualizations. Each subfigure shows the top-activating patches for a single SNS neuron across multiple images, with non-activating regions masked.}
\label{fig:additional_neuron_vis}
\end{figure}

\clearpage
\section{Additional Question-Guided Focus Examples}
\label{app:neuron_focus}

\begin{figure}[htbp]
    \centering
    \begin{subfigure}{\linewidth}
        \centering
        \includegraphics[width=\linewidth]{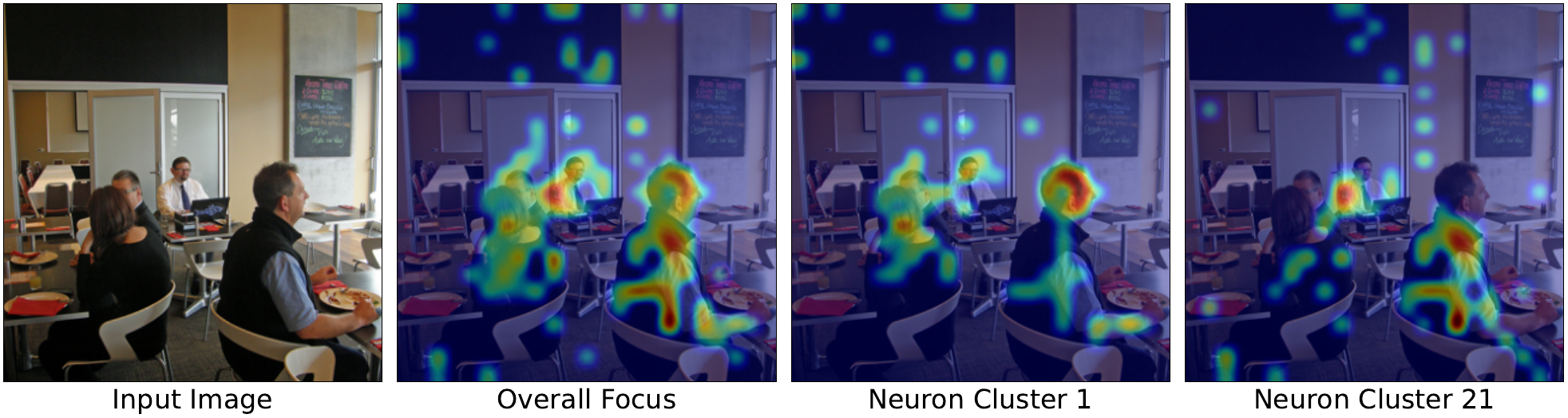}
        \caption{How many people are wearing a red shirt?}
        \label{fig:focus1}
    \end{subfigure}

    \begin{subfigure}{\linewidth}
        \centering
        \includegraphics[width=\linewidth]{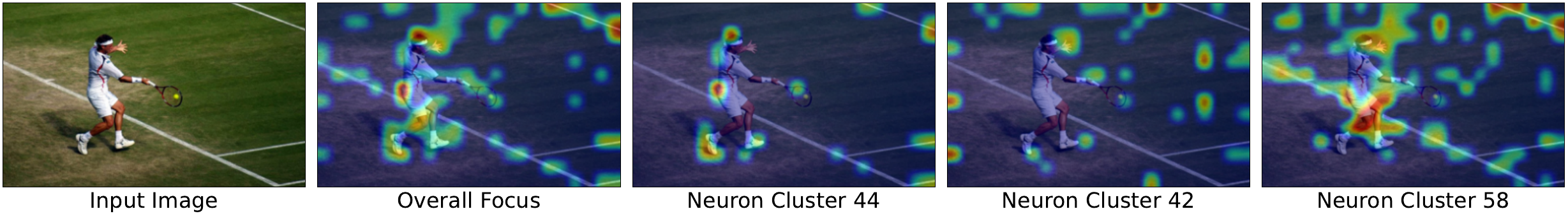}
        \caption{Did the person hit the ball?}
        \label{fig:focus3}
    \end{subfigure}

    \begin{subfigure}{\linewidth}
        \centering
        \includegraphics[width=\linewidth]{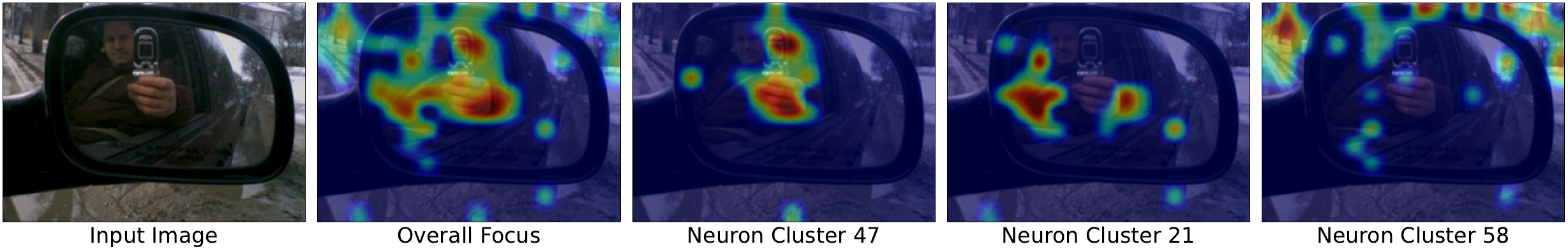}
        \caption{Where is the man looking at?}
        \label{fig:focus4}
    \end{subfigure}

    \begin{subfigure}{\linewidth}
        \centering
        \includegraphics[width=\linewidth]{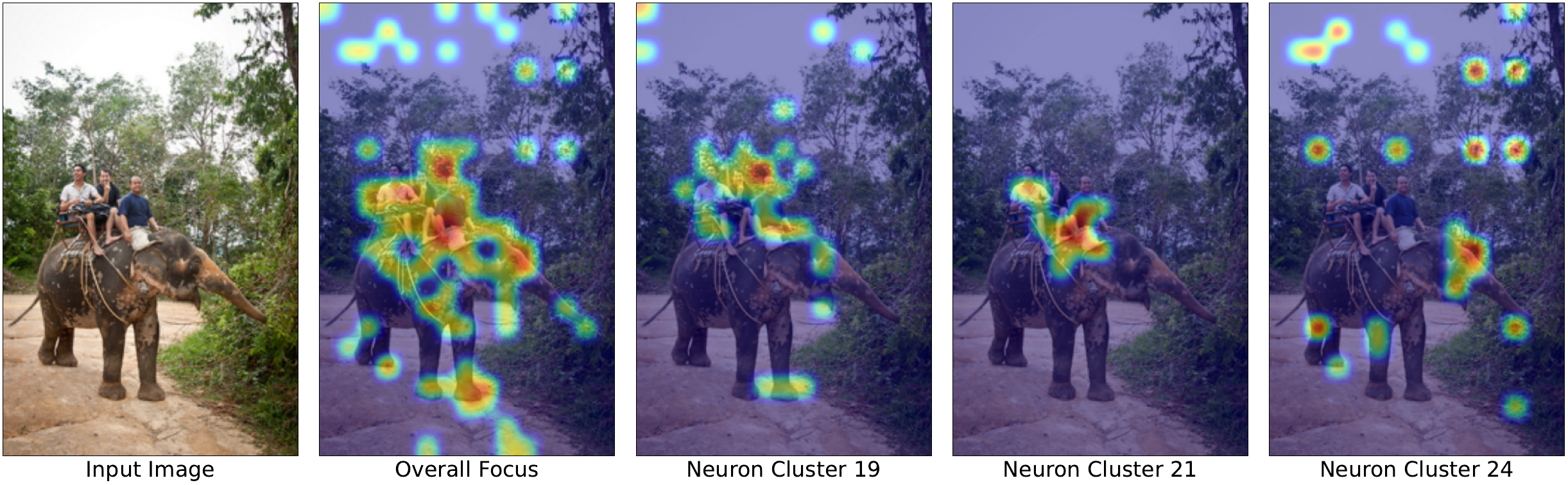}
        \caption{How many people are on the elephant?}
        \label{fig:focus5}
    \end{subfigure}

    \begin{subfigure}{\linewidth}
        \centering
        \includegraphics[width=\linewidth]{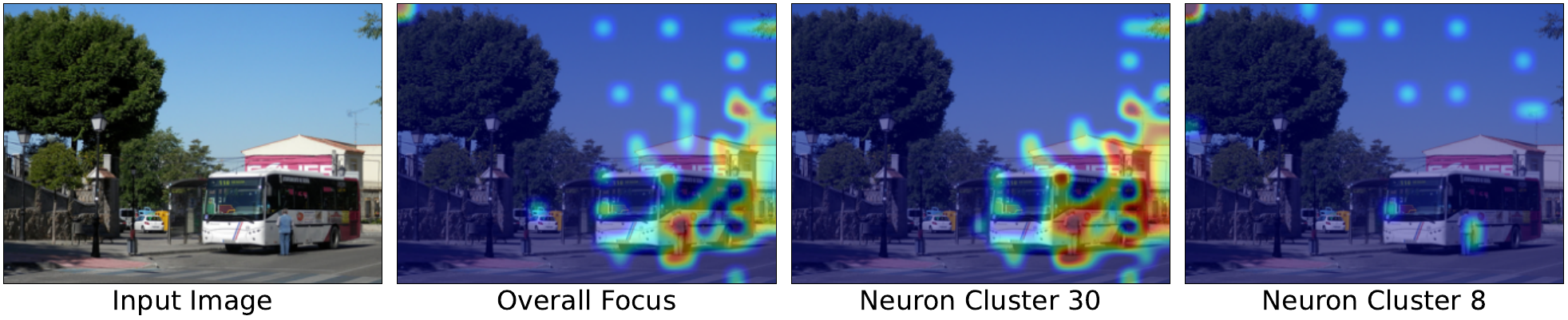}
        \caption{Are there passengers waiting to board?}
        \label{fig:focus6}
    \end{subfigure}

    \caption{Additional question-guided visual focus examples.}
    \label{fig:appendix_focus}
\end{figure}

Figure~\ref{fig:appendix_focus} provides additional examples of query-guided visual focus produced by NVF. 
In each case, NVF selects a small set of query-relevant neuron clusters and increases the activation of neurons within these clusters, yielding cluster-specific activation maps that highlight localized image regions relevant to answering the question.

\clearpage
\section{Additional Domain-Shift Evaluation}
\label{app:medical_domain_shift}

\begin{table}[htbp]
\centering
\caption{Performance comparison on OmniMedVQA-Mini by overall accuracy and question type.}
\label{tab:omnimedvqa_question_type}
\begin{tabular}{lccc}
\toprule
\textbf{Evaluation Setting} & \textbf{Base Qwen2.5-VL} & \textbf{NeuronEye} & \textbf{Difference} \\
\midrule
Overall Accuracy & 65.30\% & 63.10\% & -2.20 \\
\midrule
Anatomy Identification & 48.52\% & 48.10\% & -0.42 \\
Disease Diagnosis & 63.74\% & 60.94\% & -2.80 \\
Lesion Grading & 57.61\% & 55.43\% & -2.18 \\
Modality Recognition & 97.57\% & 96.18\% & -1.39 \\
Other Biological Attributes & 71.72\% & 66.21\% & -5.51 \\
\bottomrule
\end{tabular}
\end{table}

Table~\ref{tab:omnimedvqa_question_type} evaluates NeuronEye under medical-domain shift on OmniMedVQA-Mini. NeuronEye decreases overall accuracy from 65.30\% to 63.10\%, with larger drops on disease diagnosis, lesion grading, and other biological attributes. This result suggests that a sparse neuron vocabulary constructed from general-domain VQAv2 activations may not fully cover specialized medical visual concepts. It supports the limitation discussed in Section~\ref{sec:pcs} and motivates domain-specific SNS construction for specialized applications.

\section{Efficiency and Overhead}
\label{app:efficiency}

\begin{table}[htbp]
\centering
\small
\caption{Inference overhead relative to the base Qwen2.5-VL model.}
\label{tab:overhead}
\begin{tabular}{lcccc}
\toprule
Configuration & Latency (s) & Lat.\ $\Delta$ (\%) & Peak Mem.\ (MB) & Mem.\ $\Delta$ (\%) \\
\midrule
Base (Qwen2.5-VL)       & 0.382 & --     & 16,857 & --     \\
+ SNS + NVF              & 0.456 & +19.4  & 23,933 & +42.0  \\
+ SNS + NVF + PCS        & 0.457 & +19.7  & 23,933 & +42.0  \\
\bottomrule
\end{tabular}
\end{table}

Table~\ref{tab:overhead} summarizes the inference overhead relative to the base model. NeuronEye introduces a moderate latency increase ($\approx$20\%), mainly due to frozen SAE encoding and localized cross-attention. The memory increase mainly comes from intermediate activations, routing buffers, and the frozen SAE, rather than additional trainable components. PCS adds negligible extra latency and no additional peak memory in this setting.

\begin{table}[htbp]
\centering
\caption{Parameter breakdown of the added components. Percentages are relative to the 7.6B-parameter base model.}
\label{tab:params}
{%
\begin{tabular}{llcccc}
\toprule
\textbf{Module} & \textbf{Component} & \textbf{Params} & \textbf{\% of Base} & \textbf{Stage 1} & \textbf{Stage 2} \\
\midrule
\multirow{2}{*}{SNS}
& SAE                    & 822.08M & 10.79\% & Train  & Frozen \\
& Neuron Clusters        & --      & --      & Built  & Frozen \\
\midrule
\multirow{5}{*}{NVF}
& Router                 & 6.54M  & 0.09\% & Train & Train \\
& Vision-Side Scorer     & 6.54M  & 0.09\% & Train & Train \\
& Projection Layers      & 12.85M & 0.17\% & --    & Train \\
& Localized Cross-Attention & 51.39M & 0.67\% & --    & Train \\
& Refinement Module      & 29.84M & 0.39\% & --    & Train \\
\midrule
PCS & $\eta_{\mathrm{param}}$ & 1      & $\approx$0\% & --    & Train \\
\midrule
\multicolumn{2}{l}{\textbf{Trainable modules}} & \textbf{107.16M} & \textbf{1.41\%} & -- & -- \\
\bottomrule
\end{tabular}%
}
\end{table}

Table~\ref{tab:params} reports the parameter breakdown of the added components. The 1.41\% parameter overhead refers only to trainable modules. The SAE is trained once for constructing SNS and then kept frozen during lightweight module training and inference. Thus, the frozen SAE contributes non-trainable parameters and memory overhead, while the trainable adaptation cost remains limited to the router, vision-side scorer, projection layers, localized cross-attention, refinement module, and PCS scalar.

\clearpage


\newpage

\end{document}